%% file: multi_task_arxiv.tex
\documentclass{article}

\PassOptionsToPackage{round}{natbib}

\usepackage[final]{neurips_2018}

\input{vcl-shortcuts.tex}

\usepackage[utf8]{inputenc} 
\usepackage[T1]{fontenc}    
\usepackage{hyperref}       
\usepackage{amsfonts}       
\usepackage{microtype}      
\usepackage{caption}
\usepackage{subcaption}
\usepackage{enumitem}
\usepackage{siunitx}
\newrobustcmd*{\bftabnum}{%
  \bfseries
  \sisetup{output-decimal-marker={\textmd{.}}}%
}

\usepackage{xpatch}
\makeatletter
\AtBeginDocument{\xpatchcmd{\@thm}{\thm@headpunct{.}}{\thm@headpunct{}}{}{}}
\makeatother

\newcommand{\eqname}[1]{\tag*{#1}}

\graphicspath{{./figures/}}

\title{Multi-Task Learning as Multi-Objective Optimization}

\author{
  Ozan Sener\\
  Intel Labs\\
  \And
  Vladlen Koltun\\
  Intel Labs\\
}

\begin{document}
\maketitle

\begin{abstract}
In multi-task learning, multiple tasks are solved jointly, sharing inductive bias between them. Multi-task learning is inherently a multi-objective problem because different tasks may conflict, necessitating a trade-off. A common compromise is to optimize a proxy objective that minimizes a weighted linear combination of per-task losses. However, this workaround is only valid when the tasks do not compete, which is rarely the case. In this paper, we explicitly cast multi-task learning as multi-objective optimization, with the overall objective of finding a Pareto optimal solution. To this end, we use algorithms developed in the gradient-based multi-objective optimization literature. These algorithms are not directly applicable to large-scale learning problems since they scale poorly with the dimensionality of the gradients and the number of tasks. We therefore propose an upper bound for the multi-objective loss and show that it can be optimized efficiently. We further prove that optimizing this upper bound yields a Pareto optimal solution under realistic assumptions. We apply our method to a variety of multi-task deep learning problems including digit classification, scene understanding (joint semantic segmentation, instance segmentation, and depth estimation), and multi-label classification. Our method produces higher-performing models than recent multi-task learning formulations or per-task training.
\end{abstract}

\section{Introduction}
\label{sec:introduction}
\input{tex/introduction.tex}

\section{Related Work}
\label{sec:related}
\input{tex/related-work.tex}

\section{Multi-Task Learning as Multi-Objective Optimization}
\label{sec:method}
\input{tex/method}

\section{Experiments}
\label{sec:experiments}
\input{tex/experiments.tex}

\section{Conclusion}
\label{sec:conclusion}
\input{tex/conclusion.tex}

\input{tex/exp_figure_page}

\appendix
\section{Proof of Theorem 1}
\input{tex/appendix-proofs.tex}
\section{Additional Results on Multi-label Classification}
\input{tex/appendix-additional-results.tex}
\section{Implementation Details}
\input{tex/appendix-implementation-details.tex}

\end{document}

%% file: vcl-shortcuts.tex
\usepackage{times}
\usepackage{epsfig}
\usepackage{graphicx}
\usepackage{float}
\usepackage{wrapfig}
\usepackage{amsmath,amssymb,amsthm}
\usepackage{algorithm,algorithmicx,algpseudocode}
\usepackage{bm,xspace}
\usepackage{comment}
\usepackage{verbatim}
\usepackage{multirow}
\usepackage{balance}
\usepackage{url}
\usepackage{booktabs}
\usepackage{etoolbox,siunitx}
\usepackage{calc}
\usepackage{pifont,hologo}
\usepackage[usenames, dvipsnames]{color}
\usepackage{nicefrac}

\usepackage[super]{nth}
\usepackage{nicefrac}
\robustify\bfseries

\usepackage{dsfont}

\newtheorem{theorem}{Theorem}

\newtheorem{definition}{Definition}

\def\xx{\mathbf{x}}

\def\zz{\mathbf{z}}

\def\LL{\mathbf{L}}
\def\MM{\mathbf{M}}

\def\ZZ{\mathbf{Z}}

\def\lL{\mathcal{L}}

\def\pP{\mathcal{P}}

\def\xX{\mathcal{X}}
\def\yY{\mathcal{Y}}

\def\btheta{{\bm\theta}}

\DeclareMathOperator*{\argmin}{arg\,min}

\DeclareMathSymbol{@}{\mathord}{letters}{"3B}

\def\latex/{\LaTeX}
\def\bibtex/{\hologo{BibTeX}}


%% file: tex/introduction.tex
One of the most surprising results in statistics is Stein's paradox. \citet{Stein1956} showed that it is better
to estimate the means of three or more Gaussian random variables using samples from all of them rather than estimating them separately, even when the Gaussians are independent. Stein's paradox was an early motivation for multi-task learning (MTL) \citep{Caruana1997}, a learning paradigm in which data from multiple tasks is used with the hope to obtain superior performance over learning each task independently. Potential advantages of MTL go beyond the direct implications of Stein's paradox, since even seemingly unrelated real world tasks have strong dependencies due to the shared processes that give rise to the data. For example, although autonomous driving and object manipulation are seemingly unrelated, the underlying data is governed by the same laws of optics, material properties, and dynamics. This motivates the use of multiple tasks as an inductive bias in learning systems.

A typical MTL system is given a collection of input points and sets of targets for various tasks per point. A common way to set up the inductive bias across tasks is to design a parametrized hypothesis class that shares some parameters across tasks. Typically, these parameters are learned by solving an optimization problem that minimizes a weighted sum of the empirical risk for each task. However, the linear-combination formulation is only sensible when there is a parameter set that is effective across all tasks. In other words, minimization of a weighted sum of empirical risk is only valid if tasks are not competing, which is rarely the case. MTL with conflicting objectives requires modeling of the trade-off between tasks, which is beyond what a linear combination achieves.

An alternative objective for MTL is finding solutions that are not dominated by any others.
Such solutions are said to be Pareto optimal. In this paper, we cast the objective of MTL in terms of finding Pareto optimal solutions.

The problem of finding Pareto optimal solutions given multiple criteria is called multi-objective optimization. A variety of algorithms for multi-objective optimization exist. One such approach is the multiple-gradient descent algorithm (MGDA), which uses gradient-based optimization and provably converges to a point on the Pareto set \citep{Desideri2012}. MGDA is well-suited for multi-task learning with deep networks. It can use the gradients of each task and solve an optimization problem to decide on an update over the shared parameters. However, there are two technical problems that hinder the applicability of MGDA on a large scale. (i) The underlying optimization problem does not scale gracefully to high-dimensional gradients, which arise naturally in deep networks. (ii) The algorithm requires explicit computation of gradients per task, which results in linear scaling of the number of backward passes and roughly multiplies the training time by the number of tasks.

In this paper, we develop a Frank-Wolfe-based optimizer that scales to high-dimensional problems. Furthermore, we provide an upper bound for the MGDA optimization objective and show that it can be computed via a single backward pass without explicit task-specific gradients, thus making the computational overhead of the method negligible. We prove that using our upper bound yields a Pareto optimal solution under realistic assumptions. The result is an exact algorithm for multi-objective optimization of deep networks with negligible computational overhead.

We empirically evaluate the presented method on three different problems. First, we perform an extensive evaluation on multi-digit classification with MultiMNIST \citep{multi_mnist}. Second, we cast multi-label classification as MTL and conduct experiments with the CelebA dataset \citep{celeba}. Lastly, we apply the presented method to scene understanding; specifically, we perform joint semantic segmentation, instance segmentation, and depth estimation on the Cityscapes dataset \citep{cityscapes}. The number of tasks in our evaluation varies from 2 to 40. Our method clearly outperforms all baselines.

%% file: tex/related-work.tex

\noindent \textbf{Multi-task learning.}
We summarize the work most closely related to ours and refer the interested reader to reviews by \citet{Ruder2017} and \citet{zhou2011malsar} for additional background.
Multi-task learning (MTL) is typically conducted via hard or soft parameter sharing. In hard parameter sharing, a subset of the parameters is shared between tasks while other parameters are task-specific. In soft parameter sharing, all parameters are task-specific but they are jointly constrained via Bayesian priors \citep{Xue2007, Bakker2003} or a joint dictionary \citep{Argyriou2007, Long2015, Yang2017, Ruder2017}.  We focus on hard parameter sharing with gradient-based optimization, following the success of deep MTL in computer vision \citep{Bilen2016, Misra2016, Rudd2016, Yang2017, Kokkinos2016, Zamir2018}, natural language processing \citep{Collobert2008, Dong2015, Liu2015, Luong2015, Hashimoto2016}, speech processing \citep{Huang2013,Seltzer2013,Huang2015}, and even seemingly unrelated domains over multiple modalities \citep{Kaiser2017}.

\citet{Baxter2000} theoretically analyze the MTL problem as interaction between individual learners and a meta-algorithm. Each learner is responsible for one task and a meta-algorithm decides how the shared parameters are updated. All aforementioned MTL algorithms use weighted summation as the meta-algorithm. Meta-algorithms that go beyond weighted summation have also been explored. \citet{Cong2014} consider the case where each individual learner is based on kernel learning and utilize multi-objective optimization. \citet{Zhang2010} consider the case where each learner is a linear model and use a task affinity matrix. \citet{Zhou2011} and \citet{Bagherjeiran2005} use the assumption that tasks share a dictionary and develop an expectation-maximization-like meta-algorithm. \citet{Miranda2012} and \citet{ZhouDi2017} use swarm optimization. None of these methods apply to gradient-based learning of high-capacity models such as modern deep networks. \citet{Kendall2018} and \citet{Chen2018} propose heuristics based on uncertainty and gradient magnitudes, respectively, and apply their methods to convolutional neural networks. Another recent work uses multi-agent reinforcement learning \citep{Rosenbaum2017}.

\noindent \textbf{Multi-objective optimization.}
Multi-objective optimization addresses the problem of optimizing a set of possibly contrasting objectives. We recommend \citet{Miettinen1999} and \citet{Ehrgott2005} for surveys of this field. Of particular relevance to our work is gradient-based multi-objective optimization, as developed by \citet{Fliege2000}, \citet{Schaffler2002}, and \citet{Desideri2012}. These methods use multi-objective Karush-Kuhn-Tucker (KKT) conditions \citep{Kuhn1951} and find a descent direction that decreases all objectives. This approach was extended to stochastic gradient descent by \citet{Peitz2017} and \citet{Poirion2017}. In machine learning, these methods have been applied to multi-agent learning \citep{Ghish2013, Pirotta2016, Parisi2014}, kernel learning \citep{Cong2014}, sequential decision making \citep{Whiteson2018}, and Bayesian optimization \citep{Shah2016, Lobato2016}. Our work applies gradient-based multi-objective optimization to multi-task learning.

%% file: tex/method.tex

Consider a multi-task learning (MTL) problem over an input space $\xX$ and a collection of task spaces $\{\yY^t\}_{t \in [T]}$, such that a large dataset of i.i.d.\ data points $\{\xx_i, y_i^1,\ldots,y^T_i\}_{i \in [N]}$ is given where $T$ is the number of tasks, $N$ is the number of data points, and $y^t_i$ is the label of the $t^{\textup{th}}$ task for the $i^{\textup{th}}$ data point.\footnote{This definition can be extended to the partially-labelled case by extending $\yY^t$ with a null label.} We further consider a parametric hypothesis class per task as $f^t(\xx;\btheta^{sh}, \btheta^t): \xX \rightarrow \yY^t$, such that some parameters ($\btheta^{sh}$) are shared between tasks and some ($\btheta^t$) are task-specific. We also consider task-specific loss functions $\lL^t(\cdot,\cdot): \yY^t \times \yY^t \rightarrow {\Bbb R}^+$.

Although many hypothesis classes and loss functions have been proposed in the MTL literature, they generally yield the following empirical risk minimization formulation:
\begin{equation}
\min_{\substack{\btheta^{sh},\\ \btheta^1,\ldots,\btheta^T}} \quad \sum_{t=1}^T c^t \hat{\lL}^t(\btheta^{sh},\btheta^t)
\label{eq:linear-combination}
\end{equation}
for some static or dynamically computed weights $c^t$ per task, where $\hat{\lL}^t(\btheta^{sh},\btheta^t)$ is the empirical loss of the task $t$, defined as $\hat{\lL}^t(\btheta^{sh},\btheta^t) \triangleq \frac{1}{N} \sum_{i} \lL\big(f^t(\xx_i;\btheta^{sh},\btheta^t), y_i^t\big)$.

Although the weighted summation formulation (\ref{eq:linear-combination}) is intuitively appealing, it typically either requires an expensive grid search over various scalings or the use of a heuristic \citep{Kendall2018, Chen2018}. A basic justification for scaling is that it is not possible to define global optimality in the MTL setting. Consider two sets of solutions $\btheta$ and $\bar{\btheta}$ such that $\hat{\lL}^{t_1}(\btheta^{sh},\btheta^{t_1}) < \hat{\lL}^{t_1}(\bar{\btheta}^{sh},\bar{\btheta}^{t_1})$ and $\hat{\lL}^{t_2}(\btheta^{sh},\btheta^{t_2}) > \hat{\lL}^{t_2}(\bar{\btheta}^{sh},\bar{\btheta}^{t_2})$, for some tasks $t_1$ and $t_2$. In other words, solution $\btheta$ is better for task $t_1$ whereas $\bar{\btheta}$ is better for $t_2$. It is not possible to compare these two solutions without a pairwise importance of tasks, which is typically not available.

Alternatively, MTL can be formulated as multi-objective optimization: optimizing a collection of possibly conflicting objectives. This is the approach we take. We specify the multi-objective optimization formulation of MTL using a vector-valued loss $\LL$:
\begin{equation}
\min_{\substack{\btheta^{sh},\\ \btheta^1,\ldots,\btheta^T}} \LL(\btheta^{sh}, \btheta^1,\ldots,\btheta^T) =
\min_{\substack{\btheta^{sh},\\ \btheta^1,\ldots,\btheta^T}} \big( \hat{\lL}^1(\btheta^{sh},\btheta^1), \ldots,  \hat{\lL}^T(\btheta^{sh},\btheta^T) \big)^\intercal .
\end{equation}

The goal of multi-objective optimization is achieving Pareto optimality.

\begin{definition}[Pareto optimality for MTL] {\ }%
\begin{enumerate}[ topsep=0pt, label=\emph{(\alph*)},align=left,leftmargin=*]
\item A solution $\btheta$ dominates a solution $\bar{\btheta}$ if \mbox{$\hat{\lL}^t(\btheta^{sh},\btheta^t)  \leq \hat{\lL}^t(\bar{\btheta}^{sh},\bar{\btheta}^t)$} for all tasks $t$ and \mbox{$\LL(\btheta^{sh}, \btheta^1,\ldots,\btheta^T) \neq\LL(\bar{\btheta}^{sh}, \bar{\btheta}^1,\ldots,\bar{\btheta}^T) $}.
\item A solution $\btheta^\star$ is called Pareto optimal if there exists no solution $\btheta$ that dominates $\btheta^\star$.
\end{enumerate}
\end{definition}

The set of Pareto optimal solutions is called the Pareto set ($\pP_{\btheta}$) and its image is called the Pareto front ($\pP_{\LL} = \{ \LL(\btheta)\}_{\btheta \in \pP_{\btheta}}$). In this paper, we focus on gradient-based multi-objective optimization due to its direct relevance to gradient-based MTL.

In the rest of this section, we first summarize in Section~\ref{sec:mgda} how multi-objective optimization can be performed with gradient descent. Then, we suggest in Section~\ref{sec:optimization} a practical algorithm for performing multi-objective optimization over very large parameter spaces. Finally, in Section~\ref{sec:approximation} we propose an efficient solution for multi-objective optimization designed directly for high-capacity deep networks. Our method scales to very large models and a high number of tasks with negligible overhead.

\subsection{Multiple Gradient Descent Algorithm}
\label{sec:mgda}

As in the single-objective case, multi-objective optimization can be solved to local optimality via gradient descent. In this section, we summarize one such approach, called the multiple gradient descent algorithm (MGDA) \citep{Desideri2012}. MGDA leverages the Karush-Kuhn-Tucker (KKT) conditions, which are necessary for optimality \citep{Fliege2000,Schaffler2002,Desideri2012}. We now state the KKT conditions for both task-specific and shared parameters:
\begin{itemize}[itemsep=0pt,topsep=0pt]
\item There exist $\alpha^1,\ldots, \alpha^T \geq 0$ such that $\sum_{t=1}^T\alpha^t = 1$ and $\sum_{t=1}^T \alpha^t \nabla_{\btheta^{sh}}  \hat{\lL}^t(\btheta^{sh},\btheta^t) = 0$
\item For all tasks $t$,  $\nabla_{\btheta^{t}}  \hat{\lL}^t(\btheta^{sh},\btheta^t) = 0$
\end{itemize}
Any solution that satisfies these conditions is called a Pareto stationary point. Although every Pareto optimal point is Pareto stationary, the reverse may not be true. Consider the optimization problem
\begin{equation}
\min_{\alpha^1,\ldots,\alpha^T}  \Bigg\{  \bigg\| \sum_{t=1}^T \alpha^t \nabla_{\btheta^{sh}}  \hat{\lL}^t(\btheta^{sh},\btheta^t) \bigg\|_2^2 \bigg |  \sum_{t=1}^T \alpha^t = 1, \alpha^t \geq 0 \quad \forall t \Bigg\}
\label{eq:kkt_opt}
\end{equation}

\citet{Desideri2012} showed that either the solution to this optimization problem is $0$ and the resulting point satisfies the KKT conditions, or the solution gives a descent direction that improves all tasks. Hence, the resulting MTL algorithm would be gradient descent on the task-specific parameters followed by solving (\ref{eq:kkt_opt}) and applying the solution ($\sum_{t=1}^T\alpha^t \nabla_{\btheta^{sh}}$) as a gradient update to shared parameters. We discuss how to solve (\ref{eq:kkt_opt}) for an arbitrary model in Section~\ref{sec:optimization} and present an efficient solution when the underlying model is an encoder-decoder in Section~\ref{sec:approximation}.

\subsection{Solving the Optimization Problem}
\label{sec:optimization}

The optimization problem defined in (\ref{eq:kkt_opt}) is equivalent to finding a minimum-norm point in the convex hull of the set of input points. This problem arises naturally in computational geometry: it is equivalent to finding the closest point within a convex hull to a given query point. It has been studied extensively \citep{Makimoto1994, Wolfe1976, Sekitani1993}. Although many algorithms have been proposed, they do not apply in our setting because the assumptions they make do not hold. Algorithms proposed in the computational geometry literature address the problem of finding minimum-norm points in the convex hull of a large number of points in a low-dimensional space (typically of dimensionality 2 or 3). In our setting, the number of points is the number of tasks and is typically low; in contrast, the dimensionality is the number of shared parameters and can be in the millions. We therefore use a different approach based on convex optimization, since (\ref{eq:kkt_opt}) is a convex quadratic problem with linear constraints.

Before we tackle the general case, let's consider the case of two tasks. The optimization problem can be defined as \mbox{$\min_{\alpha \in [0,1]} \| \alpha \nabla_{\btheta^{sh}}\hat{\lL}^1(\btheta^{sh},\btheta^1)+ (1-\alpha) \nabla_{\btheta^{sh}}\hat{\lL}^2(\btheta^{sh},\btheta^2) \|_2^2$}, which is a one-dimensional quadratic function of $\alpha$ with an analytical solution:
\begin{equation}
\hat{\alpha}= \left[\frac{\big(\nabla_{\btheta^{sh}}\hat{\lL}^2(\btheta^{sh},\btheta^2) - \nabla_{\btheta^{sh}}\hat{\lL}^1(\btheta^{sh},\btheta^1)\big)^\intercal  \nabla_{\btheta^{sh}}\hat{\lL}^2(\btheta^{sh},\btheta^2)  }{\|\nabla_{\btheta^{sh}}\hat{\lL}^1(\btheta^{sh},\btheta^1) - \nabla_{\btheta^{sh}}\hat{\lL}^2(\btheta^{sh},\btheta^2)\|_2^2}\right]_{+,{1 \atop \intercal}}
\label{eq:two_task_sol}
\end{equation}
where $[\cdot]_{+,{1 \atop \intercal}}$ represents clipping to $[0,1]$ as $[a]_{+,{1 \atop \intercal}} = \max(\min(a, 1),0)$. We further visualize this solution in Figure~1. Although this is only applicable when $T=2$, this enables efficient application of the Frank-Wolfe algorithm \citep{Jaggi2013} since the line search can be solved analytically. Hence, we use Frank-Wolfe to solve the constrained optimization problem, using (\ref{eq:two_task_sol}) as a subroutine for the line search. We give all the update equations for the Frank-Wolfe solver in Algorithm~\ref{alg:mtl_mgda}.

\begin{minipage}{\textwidth}
\begin{tabular}{@{}l@{\hspace{1mm}}r@{}}
\begin{minipage}{0.68\textwidth}
\includegraphics[width=\textwidth]{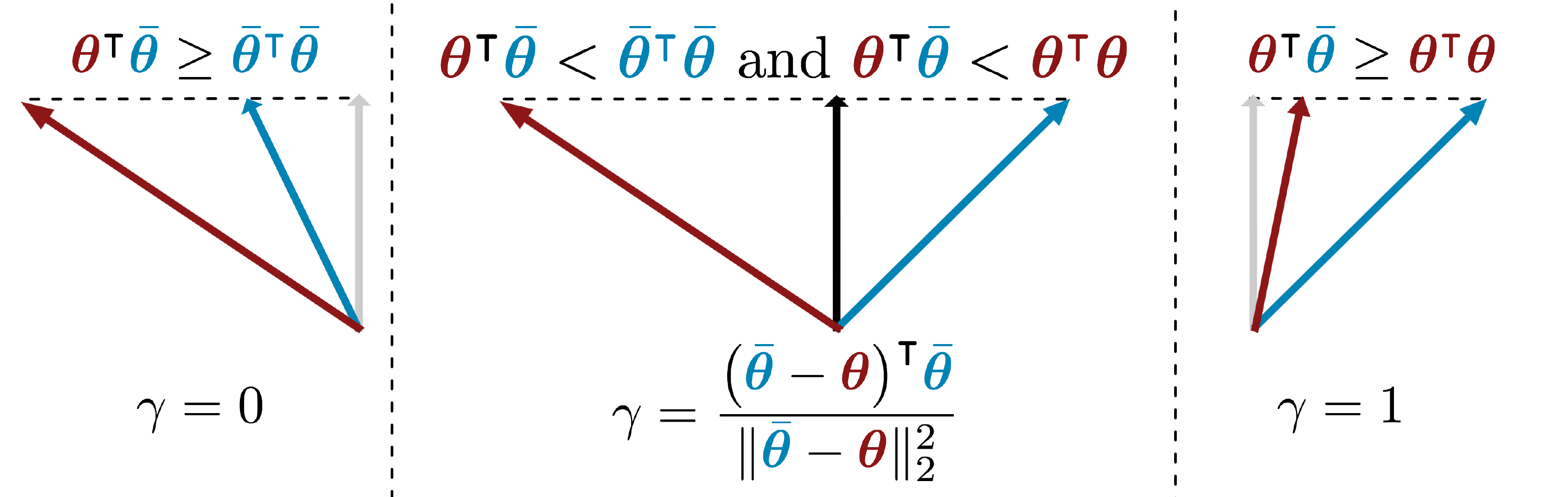}
\captionof{figure}{Visualisation of the min-norm point in the convex hull of two points (\mbox{$\min_{\gamma \in [0,1]} \| \gamma \btheta + (1-\gamma) \bar{\btheta} \|_2^2$}). As the geometry suggests, the solution is either an edge case or a perpendicular vector.}
\end{minipage}
&\begin{minipage}{0.30\textwidth}
\vspace{-5mm}
\begin{algorithm}[H]\captionsetup{labelsep=newline}
\caption{ $\min_{\gamma \in [0,1]} \| \gamma \btheta + (1-\gamma) \bar{\btheta} \|_2^2$}
\label{alg:TWO_PNT}
\begin{algorithmic}[1]
\If{$\btheta^\intercal \bar{\btheta} \geq \btheta^\intercal \btheta$}
\State $\gamma = 1$
\ElsIf{$\btheta^\intercal \bar{\btheta} \geq \bar{\btheta}^\intercal \bar{\btheta}$}
\State $\gamma = 0$
\Else
\State $\gamma = \frac{(\bar{\btheta}- \btheta)^\intercal \bar{\btheta}}{\|\btheta - \bar{\btheta}\|_2^2}$
\EndIf
\end{algorithmic}
\end{algorithm}
\end{minipage}
\end{tabular}
\begin{algorithm}[H]
\caption{Update Equations for MTL}
\label{alg:mtl_mgda}
\begin{algorithmic}[1]
\For{$t=1$ {\bfseries to} $T$}
\State $\btheta^t = \btheta^t - \eta \nabla_{\btheta^{t}}  \hat{\lL}^t(\btheta^{sh},\btheta^t)$  \Comment{Gradient descent on task-specific parameters}
\EndFor
\State $\alpha^1,\ldots,\alpha^{T}$ = \textproc{FrankWolfeSolver}($\btheta$) \Comment{Solve (3) to find a common descent direction}
\State $\btheta^{sh} = \btheta^{sh} -   \eta \sum_{t=1}^T \alpha^t \nabla_{\btheta^{sh}}  \hat{\lL}^t(\btheta^{sh},\btheta^t)$ \Comment{Gradient descent on shared parameters}
\Statex
\Procedure{FrankWolfeSolver}{$\btheta$}
\State Initialize $\bm{\alpha} = (\alpha^1, \ldots, \alpha^{T}) = (\frac{1}{T}, \ldots, \frac{1}{T})$
\State Precompute $\MM$ st. $\MM_{i,j} = \big(\nabla_{\btheta^{sh}}  \hat{\lL}^i(\btheta^{sh},\btheta^i)\big)^\intercal \big(\nabla_{\btheta^{sh}}  \hat{\lL}^j(\btheta^{sh},\btheta^j)\big)$
\Repeat
\State $\hat{t} = \argmin_r \sum_t \alpha^t \MM_{rt}$
\State $\hat{\gamma} = \argmin_{\gamma}  \big( (1 - \gamma) \bm{\alpha} + \gamma \bm{e}_{\hat{t}}  \big)^\intercal \MM  \big( (1 - \gamma) \bm{\alpha} + \gamma \bm{e}_{\hat{t}}  \big)$ \Comment{Using Algorithm 1}
\State $\bm{\alpha} = (1- \hat{\gamma})\bm{\alpha} + \hat{\gamma} \bm{e}_{\hat{t}}$
\Until{$\hat{\gamma} \sim 0$ {\bfseries or} Number of Iterations Limit}
\State {\bfseries return} $\alpha^1,\ldots,\alpha^{T}$
\EndProcedure
\end{algorithmic}
\end{algorithm}
\end{minipage}

\subsection{Efficient Optimization for Encoder-Decoder Architectures}
\label{sec:approximation}

The MTL update described in Algorithm~\ref{alg:mtl_mgda} is applicable to any problem that uses optimization based on gradient descent. Our experiments also suggest that the Frank-Wolfe solver is efficient and accurate as it typically converges in a modest number of iterations with negligible effect on training time. However, the algorithm we described needs to compute $\nabla_{\btheta^{sh}} \hat{\lL}^t(\btheta^{sh},\btheta^t)$ for each task $t$, which requires a backward pass over the shared parameters for each task. Hence, the resulting gradient computation would be the forward pass followed by $T$ backward passes. Considering the fact that computation of the backward pass is typically more expensive than the forward pass, this results in linear scaling of the training time and can be prohibitive for problems with more than a few tasks.

We now propose an efficient method that optimizes an upper bound of the objective and requires only a single backward pass. We further show that optimizing this upper bound yields a Pareto optimal solution under realistic assumptions. The architectures we address conjoin a shared representation function with task-specific decision functions. This class of architectures covers most of the existing deep MTL models and can be formally defined by constraining the hypothesis class as
\begin{equation}
f^t(\xx;\btheta^{sh},\btheta^t) = (f^t(\cdot; \btheta^t) \circ g(\cdot; \btheta^{sh})) (\xx)  = f^t( g(\xx; \btheta^{sh}); \btheta^t  )
\end{equation}
where $g$ is the representation function shared by all tasks and $f^t$ are the task-specific functions that take this representation as input. If we denote the representations as \mbox{$\ZZ = \big(\zz_1,\ldots,\zz_N \big)$}, where $\zz_i = g(\xx_i;\btheta^{sh})$, we can state the following upper bound as a direct consequence of the chain rule:
\begin{equation}
  \Bigg\| \sum_{t=1}^T \alpha^t \nabla_{\btheta^{sh}}  \hat{\lL}^t(\btheta^{sh},\btheta^t) \Bigg\|_2^2  \leq    \Bigg\|\frac{\partial \ZZ}{\partial \btheta^{sh}}\Bigg\|_2^2 \Bigg\| \sum_{t=1}^T \alpha^t  \nabla_{\ZZ}  \hat{\lL}^t(\btheta^{sh},\btheta^t) \Bigg\|_2^2
  \end{equation}
where  $\left\|\frac{\partial \ZZ}{\partial \btheta^{sh}}\right\|_2$ is the matrix norm of the Jacobian of $\ZZ$ with respect to $\btheta^{sh}$. Two desirable properties of this upper bound are that (i) $\nabla_{\ZZ} \hat{\lL}^t(\btheta^{sh},\btheta^t)$ can be computed in a single backward pass for all tasks and (ii) $\left\|\frac{\partial \ZZ}{\partial \btheta^{sh}}\right\|_2^2 $ is not a function of $\alpha^1,\ldots ,\alpha^T$, hence it can be removed when it is used as an optimization objective. We replace the $\left\| \sum_{t=1}^T \alpha^t \nabla_{\btheta^{sh}}  \hat{\lL}^t(\btheta^{sh},\btheta^t) \right\|_2^2$ term with the upper bound we have just derived in order to obtain the approximate optimization problem and drop the $\left\|\frac{\partial \ZZ}{\partial \btheta^{sh}}\right\|_2^2 $ term since it does not affect the optimization. The resulting optimization problem is
\begin{equation}
\min_{\alpha^1,\ldots,\alpha^T}  \Bigg\{  \bigg\| \sum_{t=1}^T \alpha^t  \nabla_{\ZZ}  \hat{\lL}^t(\btheta^{sh},\btheta^t) \bigg\|_2^2 \bigg |  \sum_{t=1}^T \alpha^t = 1, \alpha^t \geq 0 \quad \forall t \Bigg\} \\
\eqname{(MGDA-UB)}
\label{eq:approx}
\end{equation}

We refer to this problem as MGDA-UB (Multiple Gradient Descent Algorithm -- Upper Bound). In practice, MGDA-UB corresponds to using the gradients of the task losses with respect to the representations instead of the shared parameters. We use Algorithm~\ref{alg:mtl_mgda} with only this change as the final method.

Although MGDA-UB is an approximation of the original optimization problem, we now state a theorem that shows that our method produces a Pareto optimal solution under mild assumptions. The proof is given in the supplement.

\begin{theorem}
Assume $\frac{\partial \mathbf{Z}}{\partial \mathbf{\theta}^{sh}}$ is full-rank. If $\alpha^{1,\ldots,T}$ is the solution of MGDA-UB, one of the following is true:
\begin{enumerate}[ topsep=0pt, label=\emph{(\alph*)},align=left,leftmargin=*]
\item $\sum_{t=1}^T \alpha^t \nabla_{\btheta^{sh}}  \hat{\lL}^t(\btheta^{sh},\btheta^t)=0$ and the current parameters are Pareto stationary.
\item $\sum_{t=1}^T \alpha^t \nabla_{\btheta^{sh}}  \hat{\lL}^t(\btheta^{sh},\btheta^t)$ is a descent direction that decreases all objectives.
\end{enumerate}
\label{thm}
\end{theorem}

This result follows from the fact that as long as $\frac{\partial \ZZ}{\partial \btheta^{sh}}$ is full rank, optimizing the upper bound corresponds to minimizing the norm of the convex combination of the gradients using the Mahalonobis norm defined by $\frac{\partial \ZZ}{\partial \btheta^{sh}}^\intercal \frac{\partial \ZZ}{\partial \btheta^{sh}}$. The non-singularity assumption is reasonable as singularity implies that tasks are linearly related and a trade-off is not necessary. In summary, our method provably finds a Pareto stationary point with negligible computational overhead and can be applied to any deep multi-objective problem with an encoder-decoder model.

%% file: tex/experiments.tex

We evaluate the presented MTL method on a number of problems. First, we use MultiMNIST \citep{multi_mnist}, an MTL adaptation of MNIST \citep{mnist}. Next, we tackle multi-label classification on the CelebA dataset \citep{celeba} by considering each label as a distinct binary classification task. These problems include both classification and regression, with the number of tasks ranging from 2 to 40. Finally, we experiment with scene understanding, jointly tackling the tasks of semantic segmentation, instance segmentation, and depth estimation on the Cityscapes dataset \citep{cityscapes}. We discuss each experiment separately in the following subsections.

The baselines we consider are (i) \textbf{uniform scaling:} minimizing a uniformly weighted sum of loss functions \mbox{$\frac{1}{T}\sum_t \lL^t$}, \mbox{(ii) \textbf{single task:}} solving tasks independently, \mbox{(iii) \textbf{grid search:}} exhaustively trying various values from $\{ c^t \in [0,1] | \sum_t c^t = 1\}$ and optimizing for $\frac{1}{T}\sum_t c^t \lL^t$, \mbox{(iv) \textbf{\citet{Kendall2018}:}} using the uncertainty weighting proposed by \citet{Kendall2018}, and \mbox{(v) \textbf{GradNorm:}} using the normalization proposed by \citet{Chen2018}.

\subsection{MultiMNIST}
\label{sec:multi_mnist_exp}

Our initial experiments are on MultiMNIST, an MTL version of the MNIST dataset \citep{multi_mnist}. In order to convert digit classification into a multi-task problem, \citet{multi_mnist} overlaid multiple images together. We use a similar construction. For each image, a different one is chosen uniformly in random. Then one of these images is put at the top-left and the other one is at the bottom-right. The resulting tasks are: classifying the digit on the top-left (task-L) and classifying the digit on the bottom-right (task-R). We use 60K examples and directly apply existing single-task MNIST models. The MultiMNIST dataset is illustrated in the supplement.

We use the LeNet architecture \citep{mnist}. We treat all layers except the last as the representation function $g$ and put two fully-connected layers as task-specific functions (see the supplement for details). We visualize the performance profile as a scatter plot of accuracies on task-L and task-R in Figure~\ref{fig:multi_mnist_performance_curve}, and list the results in Table~\ref{tab:multi_mnist}.

In this setup, any static scaling results in lower accuracy than solving each task separately (the single-task baseline). The two tasks appear to compete for model capacity, since increase in the accuracy of one task results in decrease in the accuracy of the other. Uncertainty weighting \citep{Kendall2018} and GradNorm \citep{Chen2018} find solutions that are slightly better than grid search but distinctly worse than the single-task baseline. In contrast, our method finds a solution that efficiently utilizes the model capacity and yields accuracies that are as good as the single-task solutions. This experiment demonstrates the effectiveness of our method as well as the necessity of treating MTL as multi-objective optimization. Even after a large hyper-parameter search, \emph{any} scaling of tasks does not approach the effectiveness of our method.

\subsection{Multi-Label Classification}

\begin{figure}[t]
\includegraphics[width=\textwidth]{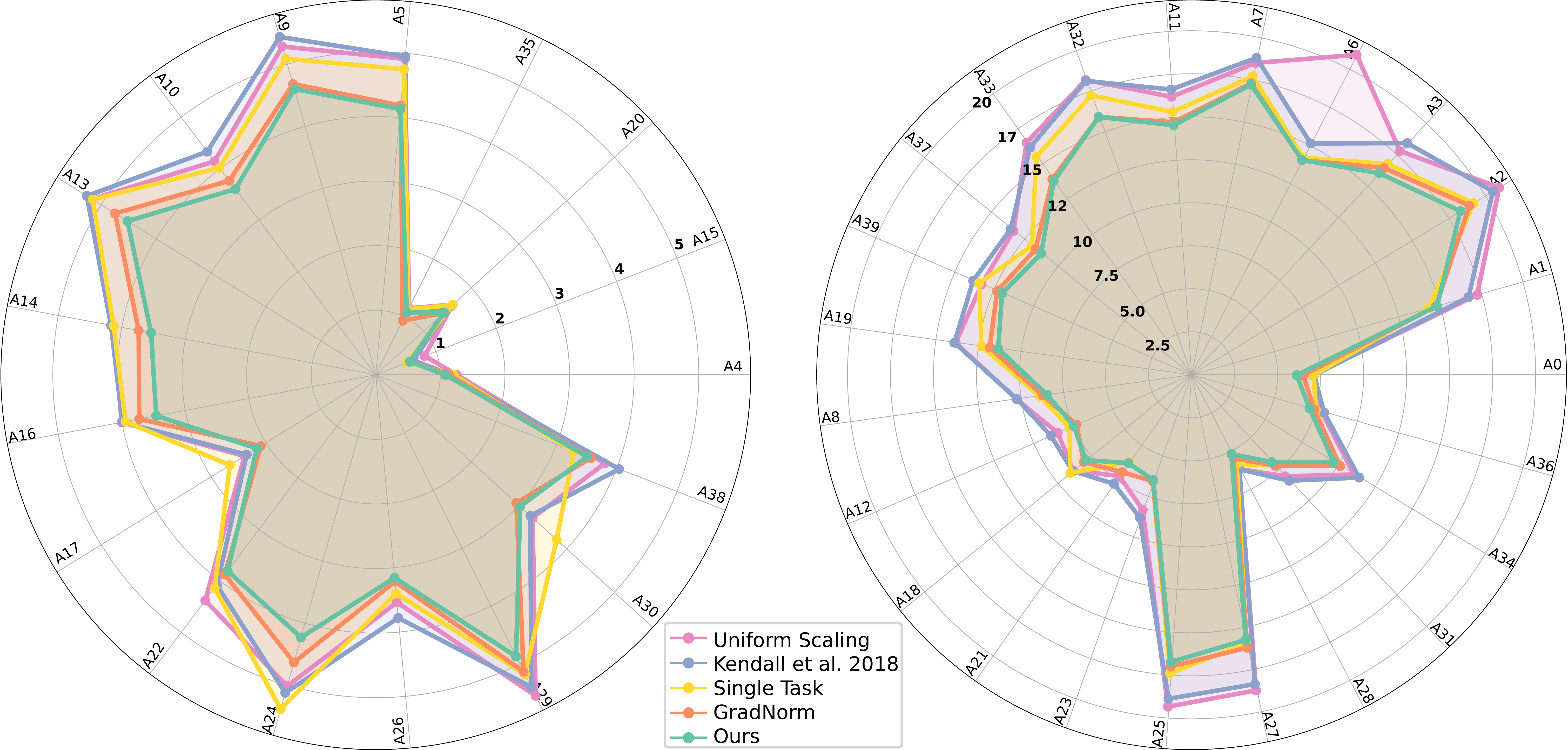}
\vspace{1mm}
\caption{Radar charts of percentage error per attribute on CelebA \citep{celeba}. Lower is better. We divide attributes into two sets for legibility: easy on the left, hard on the right. Zoom in for details.}
\label{fig:multi_label_radar}
\end{figure}

\begin{wraptable}{r}{0.3\textwidth}
\captionof{table}{Mean of error per category of MTL algorithms in multi-label classification on CelebA \citep{celeba}.}
\begin{tabular}{r@{\hspace{2mm}}c@{}}
\toprule
& Average  \\
&  error \\
\midrule
Single task & $8.77$ \\
Uniform scaling & $9.62$ \\
\citealt{Kendall2018} & $9.53$ \\
GradNorm & $8.44$ \\
Ours & $\mathbf{8.25}$  \\
\bottomrule
\end{tabular}
\label{table:multi_label_bar}
\end{wraptable}

Next, we tackle multi-label classification. Given a set of attributes, multi-label classification calls for deciding whether each attribute holds for the input. We use the CelebA dataset \citep{celeba}, which includes 200K face images annotated with 40 attributes. Each attribute gives rise to a binary classification task and we cast this as a 40-way MTL problem. We use ResNet-18 \citep{resnet} without the final layer as a shared representation function, and attach a linear layer for each attribute (see the supplement for further details).

We plot the resulting error for each binary classification task as a radar chart in Figure~\ref{fig:multi_label_radar}. The average over them is listed in Table~\ref{table:multi_label_bar}. We skip grid search since it is not feasible over 40 tasks. Although uniform scaling is the norm in the multi-label classification literature, single-task performance is significantly better. Our method outperforms baselines for significant majority of tasks and achieves comparable performance in rest. This experiment also shows that our method remains effective when the number of tasks is high.

\subsection{Scene Understanding}

To evaluate our method in a more realistic setting, we use scene understanding. Given an RGB image, we solve three tasks: semantic segmentation (assigning pixel-level class labels), instance segmentation (assigning pixel-level instance labels), and monocular depth estimation (estimating continuous disparity per pixel). We follow the experimental procedure of \citet{Kendall2018} and use an encoder-decoder architecture. The encoder is based on ResNet-50 \citep{resnet} and is shared by all three tasks. The decoders are task-specific and are based on the pyramid pooling module \citep{pspnet} (see the supplement for further implementation details).

Since the output space of instance segmentation is unconstrained (the number of instances is not known in advance), we use a proxy problem as in \citet{Kendall2018}. For each pixel, we estimate the location of the center of mass of the instance that encompasses the pixel. These center votes can then be clustered to extract the instances. In our experiments, we directly report the MSE in the proxy task. Figure~\ref{fig:cityscapes_performance_profile} shows the performance profile for each pair of tasks, although we perform all experiments on all three tasks jointly. The pairwise performance profiles shown in Figure~\ref{fig:cityscapes_performance_profile} are simply 2D projections of the three-dimensional profile, presented this way for legibility. The results are also listed in Table~\ref{tab:cityscapes_results}.

MTL outperforms single-task accuracy, indicating that the tasks cooperate and help each other. Our method outperforms all baselines on all tasks.

\subsection{Role of the Approximation}

In order to understand the role of the approximation proposed in Section~\ref{sec:approximation}, we compare the final performance and training time of our algorithm with and without the presented approximation in Table~\ref{tab:approximation_tradeoff} (runtime measured on a single Titan Xp GPU). For a small number of tasks (3 for scene understanding), training time is reduced by 40\%. For the multi-label classification experiment (40 tasks), the presented approximation accelerates learning by a factor of 25.

On the accuracy side, we expect both methods to perform similarly as long as the full-rank assumption is satisfied. As expected, the accuracy of both methods is very similar. Somewhat surprisingly, our approximation results in slightly improved accuracy in all experiments. While counter-intuitive at first, we hypothesize that this is related to the use of SGD in the learning algorithm. Stability analysis in convex optimization suggests that if gradients are computed with an error $\hat{\nabla}_\btheta \mathcal{L}^t = \nabla_\btheta \mathcal{L}^t + \mathbf{e}^t$ ($\btheta$ corresponds to $\btheta^{sh}$ in (\ref{eq:kkt_opt})), as opposed to $\mathbf{Z}$ in the approximate problem in \ref{eq:approx}, the error in the solution is bounded as $\|\hat{\mathbf{\alpha}} - \mathbf{\alpha} \|_2 \leq \mathcal{O}(\max_t \|\mathbf{e}^t\|_2)$. Considering the fact that the gradients are computed over the full parameter set (millions of dimensions) for the original problem and over a smaller space for the approximation (batch size times representation which is in the thousands), the dimension of the error vector is significantly higher in the original problem. We expect the $l_2$ norm of such a random vector to depend on the dimension.

In summary, our quantitative analysis of the approximation suggests that (i) the approximation does not cause an accuracy drop and (ii) by solving an equivalent problem in a lower-dimensional space, our method achieves both better computational efficiency and higher stability.

  {\small
  \begin{table}[t]
  \caption{Effect of the MGDA-UB approximation. We report the final accuracies as well as training times for our method with and without the approximation.}
  \centering
  \begin{tabular}{@{}r@{\hspace{3mm}}c@{\hspace{3mm}}c@{\hspace{2mm}}c@{\hspace{2mm}}c@{}c@{\hspace{5mm}}c@{\hspace{2mm}}c@{}}
  \toprule
  & \multicolumn{4}{c}{Scene understanding (3 tasks)} &  & \multicolumn{2}{c}{Multi-label (40 tasks)}  \\
  \cmidrule(r){2-5} \cmidrule(lr){7-8}
                  & Training & Segmentation & Instance  & Disparity      & & Training & Average \\
                 & time     &  mIoU [\%]       & error [px] & error [px] & & time (hour)      & error \\
  \midrule
  Ours (w/o approx.) & $38.6$ & $66.13$ & $10.28$ & $2.59$ & & $429.9$ & $8.33$ \\
  Ours & $\mathbf{23.3}$ & $\mathbf{66.63}$ & $\mathbf{10.25}$ & $\mathbf{2.54}$  & & $\mathbf{16.1}$ & $\mathbf{8.25}$ \\
  \bottomrule
  \end{tabular}
  \label{tab:approximation_tradeoff}
  \end{table}}

%% file: tex/conclusion.tex

We described an approach to multi-task learning. Our approach is based on multi-objective optimization. In order to apply multi-objective optimization to MTL, we described an efficient algorithm as well as specific approximations that yielded a deep MTL algorithm with almost no computational overhead. Our experiments indicate that the resulting algorithm is effective for a wide range of multi-task scenarios.

%% file: tex/exp_figure_page.tex
\clearpage

\begin{minipage}{\textwidth}
\begin{tabular}{@{}c@{\hspace{11mm}}c@{}}
\begin{minipage}{.475\textwidth}
  \centering
\includegraphics[width=\textwidth]{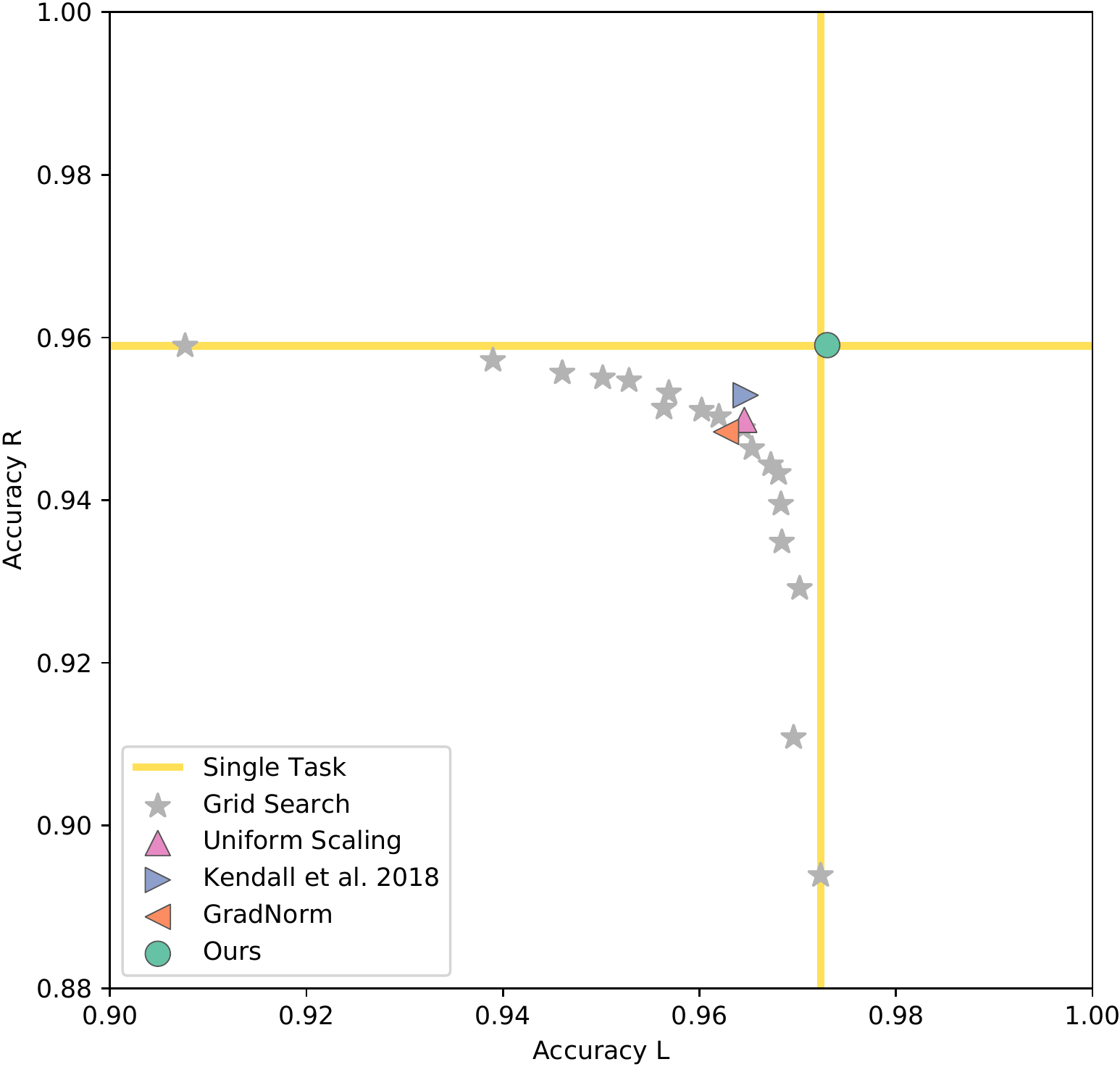}
  \captionof{figure}{\textbf{MultiMNIST accuracy profile.} We plot the obtained accuracy in detecting the left and right digits for all baselines. The grid-search results suggest that the tasks compete for model capacity. Our method is the only one that finds a solution that is as good as training a dedicated model for each task. Top-right is better.}
  \label{fig:multi_mnist_performance_curve}
  \vspace{7mm}
  \captionof{table}{Performance of MTL algorithms on MultiMNIST. Single-task baselines solve tasks separately, with dedicated models, but are shown in the same row for clarity.}
\resizebox{\textwidth}{!}{%
\begin{tabular}{r@{\hspace{4mm}}c@{\hspace{4mm}}c}
\toprule
& Left digit & Right digit   \\
& accuracy [\%] & accuracy [\%]   \\
\midrule
Single task & $\mathbf{97.23}$ & $\mathbf{95.90}$ \\
Uniform scaling & $96.46$ & $94.99$ \\
\citealt{Kendall2018} & $96.47$ & $95.29$ \\
GradNorm & $96.27$ & $94.84$ \\
Ours & $\mathbf{97.26}$ & $\mathbf{95.90}$ \\
\bottomrule
\end{tabular}}
\label{tab:multi_mnist}
  \vspace{7mm}
\captionof{table}{Performance of MTL algorithms in joint semantic segmentation, instance segmentation, and depth estimation on Cityscapes. Single-task baselines solve tasks separately but are shown in the same row for clarity.}
\resizebox{\textwidth}{!}{%
\begin{tabular}{r@{\hspace{1mm}}c@{\hspace{2mm}}c@{\hspace{2mm}}c@{}}
\toprule
& Segmentation & Instance & Disparity  \\
& mIoU [\%] &  error [px] &  error [px]  \\
\midrule
Single task & $60.68$ & $11.34$ &  $2.78$\\
Uniform scaling & $54.59$ & $10.38$ & $2.96$ \\
\citealt{Kendall2018} & $64.21$ & $11.54$ & $2.65$ \\
 GradNorm  & $64.81$ & $11.31$ & $\mathbf{2.57}$ \\
Ours & $\mathbf{66.63}$ & $\mathbf{10.25}$ & $\mathbf{2.54}$  \\
\bottomrule
\end{tabular}}
\label{tab:cityscapes_results}
\end{minipage}&
\begin{minipage}{.475\textwidth}
  \centering
  \begin{tabular}{@{}r@{}}
  \vspace{1mm} \\
  \includegraphics[width=\textwidth]{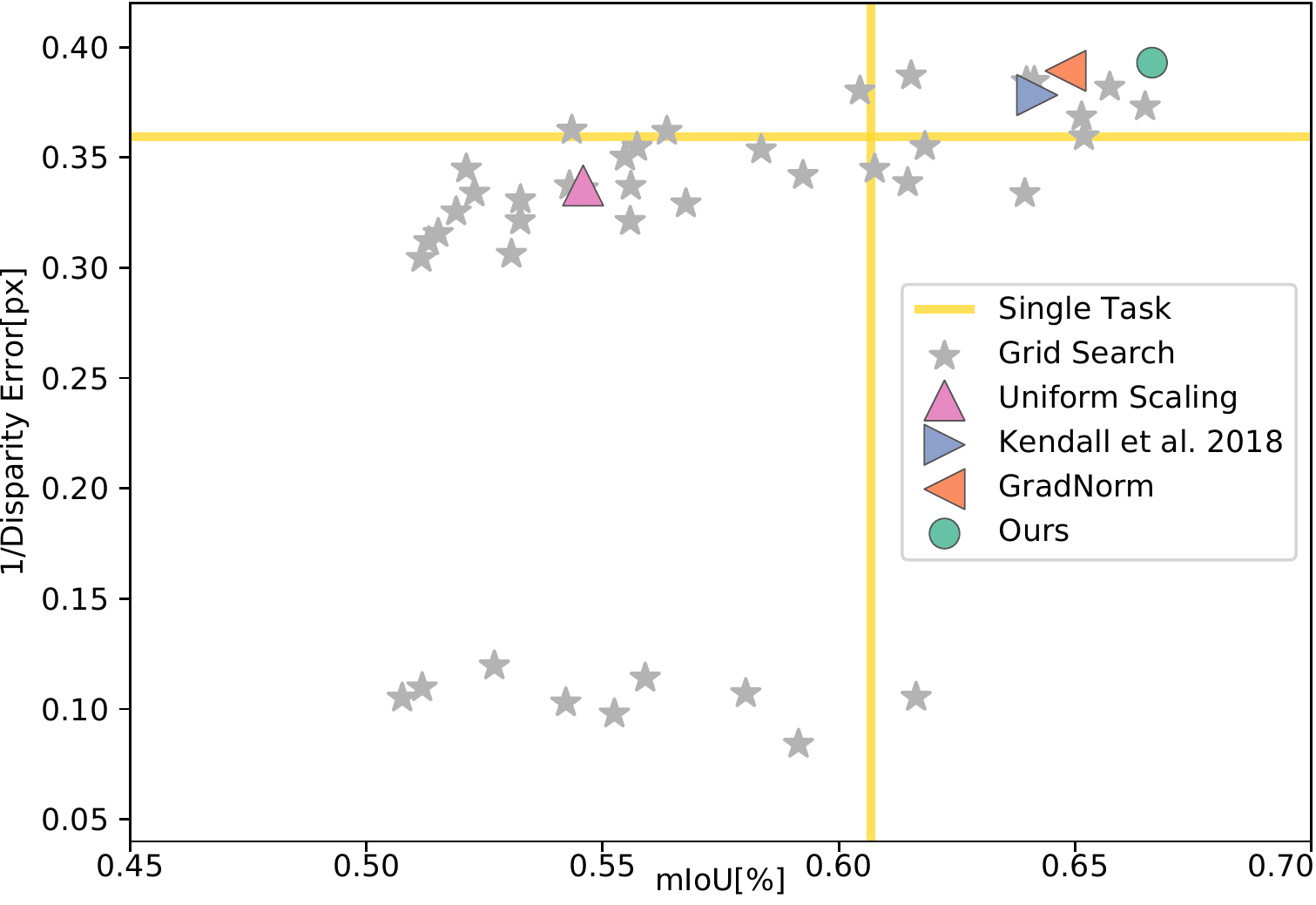}   \\ \vspace{2mm} \\
 \includegraphics[width=\textwidth]{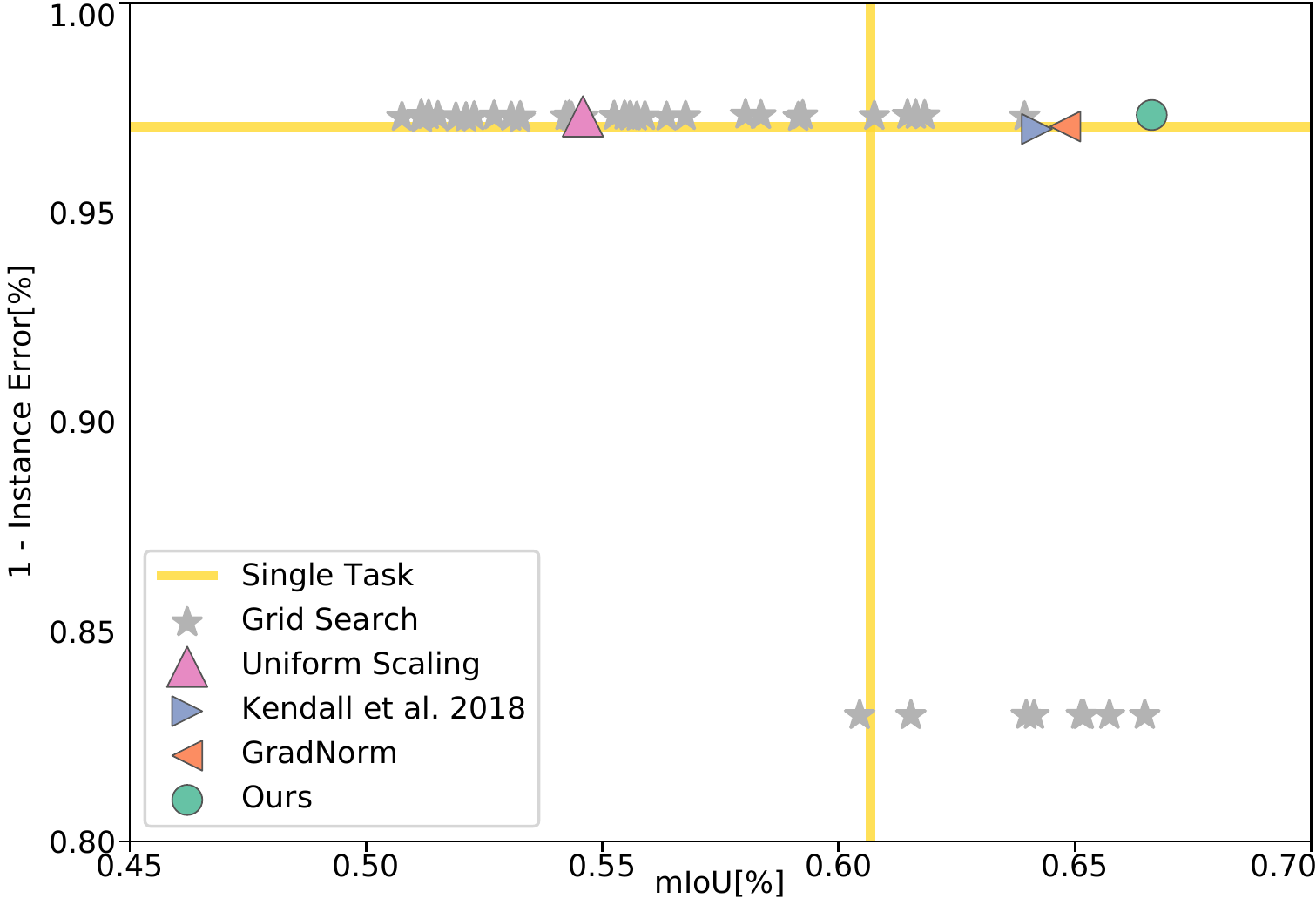}  \\ \vspace{2mm} \\
 \includegraphics[width=\textwidth]{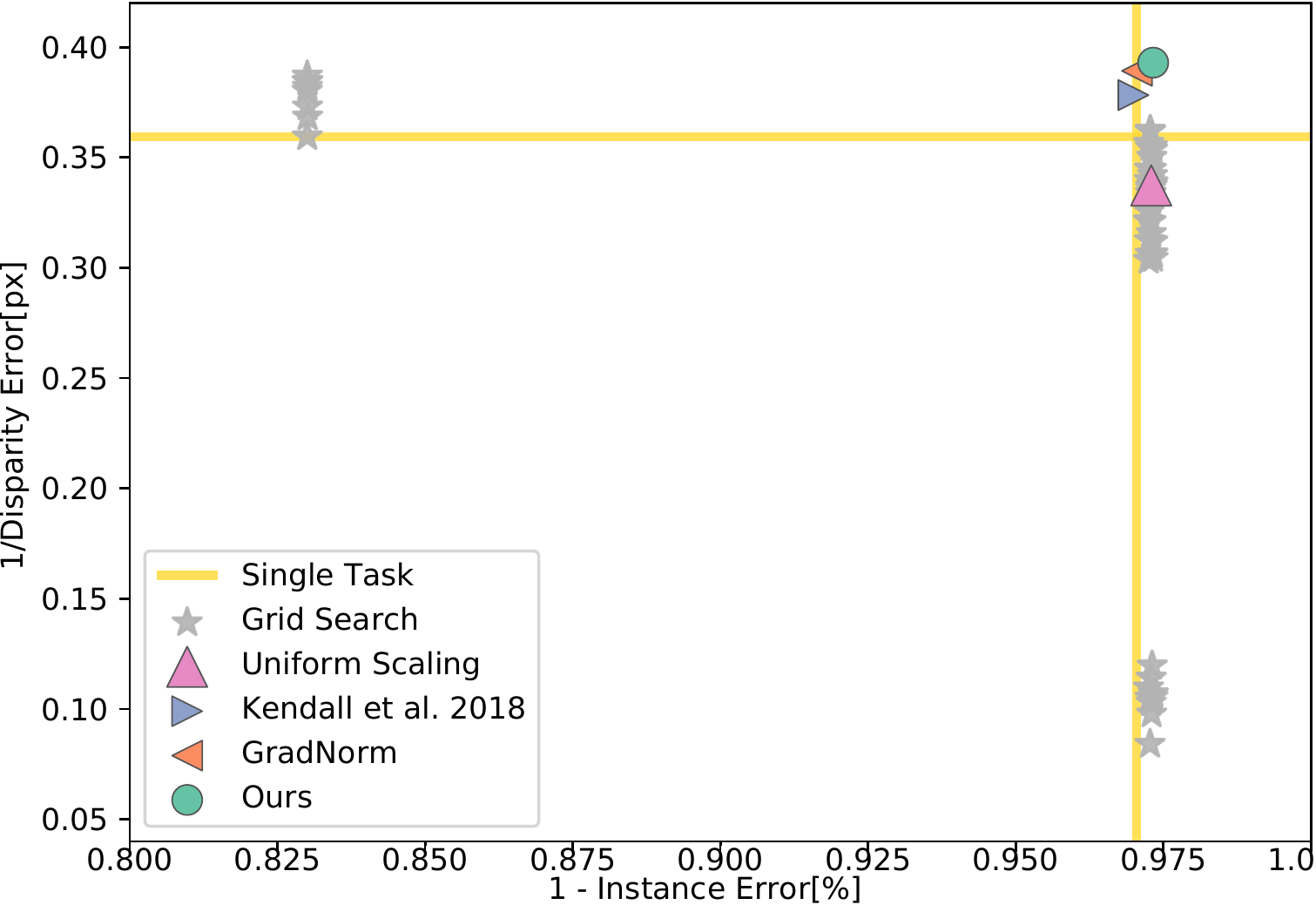}
\end{tabular}
  \captionof{figure}{\textbf{Cityscapes performance profile.} We plot the performance of all baselines for the tasks of semantic segmentation, instance segmentation, and depth estimation. We use mIoU for semantic segmentation, error of per-pixel regression (normalized to image size) for instance segmentation, and disparity error for depth estimation. To convert errors to performance measures, we use 1 $-$ instance error and 1/disparity error. We plot 2D projections of the performance profile for each pair of tasks. Although we plot pairwise projections for visualization, each point in the plots solves all tasks. Top-right is better.}
  \label{fig:cityscapes_performance_profile}
\end{minipage}%
\end{tabular}
\end{minipage}

%% file: tex/appendix-proofs.tex

\begin{proof}
We begin by showing that if the optimum value of \ref{eq:approx} is 0, so is the optimum value of (\ref{eq:kkt_opt}). This shows the first case of the theorem. Then, we will show the second part.

If the optimum value of \ref{eq:approx} is $0$,
\begin{equation}
\sum_{t=1}^T \alpha^t \nabla_{\btheta^{sh}}  \hat{\lL}^t(\btheta^{sh},\btheta^t) = \frac{\partial \mathbf{Z}}{\partial \mathbf{\theta}^{sh}} \sum_{t=1}^T \alpha^t \nabla_{\mathbf{Z}} \hat{\mathcal{L}^t} = \sum_{t=1}^T \alpha^t \nabla_{\mathbf{\theta}^{sh}} \hat{\mathcal{L}^t}=0
\end{equation}

Hence $\alpha^1,\ldots,\alpha^T$ is the solution of (\ref{eq:kkt_opt}) and the optimal value of (\ref{eq:kkt_opt}) is $0$. This proves the first case of the theorem. Before we move to the second case, we state a straightforward corollary. Since $ \frac{\partial \mathbf{Z}}{\partial \mathbf{\theta}^{sh}}$ is  full rank, this equivalence is bi-directional. In other words, if $\alpha^1,\ldots,\alpha^T$ is the solution of (\ref{eq:kkt_opt}), it is the solution of \ref{eq:approx} as well. Hence, both formulations completely agree on Pareto stationarity.

In order to prove the second case, we need to show that the resulting descent direction computed by solving \ref{eq:approx} does not increase any of the loss functions. Formally, we need to show that

\begin{equation}
\left(\sum_{t=1}^T \alpha^t \nabla_{\mathbf{\theta}^{sh}} \hat{\mathcal{L}}^t\right)^\intercal \left( \nabla_{\mathbf{\theta}^{sh}} \hat{\mathcal{L}}^{t^\prime}  \right) \geq 0 \quad \forall ~t^\prime \in \{1, \ldots, T\}
\end{equation}

This condition is equivalent to
\begin{equation}
\left(\sum_{t=1}^T \alpha^t \nabla_{\mathbf{Z}} \hat{\mathcal{L}}^t\right)^\intercal \mathbf{M} \left(\nabla_{\mathbf{Z}} \hat{\mathcal{L}}^{t^\prime}\right) \geq 0 \quad \forall ~ t^\prime \in \{1, \ldots, T\}
\end{equation}
where $\mathbf{M}=\big(\frac{\partial \mathbf{Z}}{\partial \mathbf{\theta}^{sh}}\big)^\intercal\big(\frac{\partial \mathbf{Z}}{\partial \mathbf{\theta}^{sh}}\big)$. Since $\mathbf{M}$ is positive definite (following the assumption), this is further equivalent to
\begin{equation}
\left(\sum_{t=1}^T \alpha^t \nabla_{\mathbf{Z}} \hat{\mathcal{L}}^t\right)^\intercal  \left(\nabla_{\mathbf{Z}} \hat{\mathcal{L}}^{t^\prime}\right) \geq 0 \quad \forall ~ t^\prime \in \{1, \ldots, T\}
\end{equation}

We show that this follows from the optimality conditions for \ref{eq:approx}. The Lagrangian of \ref{eq:approx} is
\begin{equation}
\left(\sum_{t=1}^T \alpha^t \nabla_{\mathbf{Z}} \hat{\mathcal{L}}^t\right)^\intercal\left(\sum_{t=1}^T \alpha^t \nabla_{\mathbf{Z}} \hat{\mathcal{L}}^t\right) - \lambda \left(\sum_i \alpha^i - 1\right) \text{ where } \lambda\geq 0.
\end{equation}

The KKT condition for this Lagrangian yields the desired result as
\begin{equation}
 \left(\sum_{t=1}^T \alpha^t \nabla_{\mathbf{Z}} \hat{\mathcal{L}}^t\right)^\intercal \left( \nabla_{\mathbf{Z}} \hat{\mathcal{L}}^t \right) = \frac{\lambda}{2} \geq 0
 \end{equation}
\end{proof}

%% file: tex/appendix-additional-results.tex

In this section, we present the experimental results we did not include in the main text. 

\label{sec:multi_label_error}
In the main text, we plotted a radar chart of the binary attribute classification errors. However, we did not include the tabulated results due to the space limitations. Here we list the binary classification error of each attribute for each algorithm in Table~\ref{tab:multi_label_table}.

\begin{table}[ht]
\caption{Multi-label classification error per attribute for all algorithms.}
\newcolumntype{Z}{S[table-format=2.2,table-auto-round]}
\resizebox{\textwidth}{!}{%
\begin{tabular}{l@{\hspace{2mm}}ZZ@{\hspace{2mm}}ZZ@{\hspace{2mm}}Z@{}c@{\hspace{10mm}}l@{\hspace{2mm}}ZZ@{\hspace{2mm}}ZZ@{\hspace{2mm}}Z}
\toprule
 & \multicolumn{1}{c}{Uniform}  &  \multicolumn{1}{c}{Single}  &  \multicolumn{1}{c}{Kendall } &      \multicolumn{1}{c}{Grad} &              &         &        &  \multicolumn{1}{c}{Uniform}  &  \multicolumn{1}{c}{Single}  &  \multicolumn{1}{c}{Kendall}  &  \multicolumn{1}{c}{Grad}& \\
   &  \multicolumn{1}{c}{scaling}  &   \multicolumn{1}{c}{task}    &   \multicolumn{1}{c}{et al.}    &   \multicolumn{1}{c}{Norm}            &  \multicolumn{1}{c}{Ours} &   &  & \multicolumn{1}{c}{scaling} &   \multicolumn{1}{c}{task} &   \multicolumn{1}{c}{et al.}   &   \multicolumn{1}{c}{Norm}   & \multicolumn{1}{c}{Ours} \\
\midrule \\
Attr. 0 & 7.11 & 7.16 & 7.18 & 6.54 & \bftabnum 6.17 & & Attr. 5 & 4.91 & 4.75 & 4.95 & 4.19 & \bftabnum 4.13 \\
Attr. 1 & 17.30 & \bftabnum 14.38 & 16.77 &  14.80 & 14.87 & & Attr. 6 & 20.97 & 14.24 & 15.17 & \bftabnum 14.07 &  14.08 \\
Attr. 2 & 20.99 & 19.25 & 20.56 & 18.97 & \bftabnum 18.35 & & Attr. 7 & 18.53 & 17.74 & 18.84 & 17.33 & \bftabnum 17.25 \\
Attr. 3 & 17.82 & 16.79 & 18.45 & 16.47 & \bftabnum 16.06 & & Attr. 8 & 10.22 & 8.87 & 10.19 & 8.67 & \bftabnum 8.42 \\
Attr. 4 & 1.25 & 1.20 & 1.17 & 1.13 & \bftabnum 1.08 & & Attr. 9 & 5.29 & 5.09 & 5.44 & 4.68 & \bftabnum 4.60 \\
\midrule
Attr. 10 & 4.14 & 4.02 & 4.33 & 3.77 & \bftabnum 3.60 & & Attr. 15 & 0.81 & \bftabnum 0.52 & 0.62 & 0.56 & 0.56 \\
Attr. 11 & 16.22 & 15.34 & 16.64 & 14.73 & \bftabnum 14.56 & & Attr. 16 & 4.00 & 3.94 & 3.99 & 3.72 & \bftabnum 3.46 \\
Attr. 12 & 8.42 & 7.68 & 8.85 & \bftabnum 7.23 &  7.41 & & Attr. 17 & 2.39 & 2.66 & 2.35 & \bftabnum 2.09 &  2.16 \\
Attr. 13 & 5.17 & 5.15 & 5.26 & 4.75 & \bftabnum 4.52 & & Attr. 18 & 8.79 & 9.01 & 8.84 & 8.00 & \bftabnum 7.83 \\
Attr. 14 & 4.14 & 4.13 & 4.17 & 3.73 & \bftabnum 3.54 & & Attr. 19 & 13.78 & 12.27 & 13.86 & 11.79 & \bftabnum 11.29 \\
\midrule
Attr. 20 & 1.61 & 1.61 & 1.58 & \bftabnum 1.42 &  1.43 & & Attr. 25 & 27.59 & 24.82 & 26.94 & 24.26 & \bftabnum 23.87 \\
Attr. 21 & 7.18 & \bftabnum 6.20 & 7.73 & 6.91 & 6.26 & & Attr. 26 & 3.54 & 3.40 & 3.78 & 3.22 & \bftabnum 3.16 \\
Attr. 22 & 4.38 & 4.14 & 4.08 & 3.88 & \bftabnum 3.81 & & Attr. 27 & 26.74 & 22.74 & 26.21 & 23.12 & \bftabnum 22.45 \\
Attr. 23 & 8.32 & 6.57 & 8.80 & 6.54 & \bftabnum 6.47 & & Attr. 28 & 6.14 & 5.82 & 6.17 & 5.43 & \bftabnum 5.16 \\
Attr. 24 & 5.01 & 5.38 & 5.12 & 4.63 & \bftabnum 4.23 & & Attr. 29 & 5.55 & 5.18 & 5.40 & 5.13 & \bftabnum 4.87 \\
\midrule
Attr. 30 & 3.29 & 3.79 & 3.24 & \bftabnum 2.94 &  3.03 & & Attr. 35 & 1.15 & 1.13 & 1.08 & \bftabnum 0.94 &  1.08 \\
Attr. 31 & 8.05 & 7.18 & 8.40 & 7.21 & \bftabnum 6.92 & & Attr. 36 & 7.91 & 7.56 & 8.06 & 7.47 & \bftabnum 7.18 \\
Attr. 32 & 18.21 & 17.25 & 18.15 & \bftabnum15.93 & \bftabnum 15.93 & & Attr. 37 & 13.27 & 11.90 & 13.47 & 11.61 & \bftabnum 11.19 \\
Attr. 33 & 16.53 & 15.55 & 16.19 & 13.93 & \bftabnum 13.80 & & Attr. 38 & 3.80 & \bftabnum 3.29 & 4.04 & 3.57 & 3.51 \\
Attr. 34 & 11.12 & 9.76 & 11.46 & 10.17 & \bftabnum 9.73 & & Attr. 39 & 13.25 & 13.40 & 13.78 & 12.26 & \bftabnum 11.95 \\
\bottomrule
\end{tabular}}
\label{tab:multi_label_table}
\end{table}

%% file: tex/appendix-implementation-details.tex

\subsection{MultiMNIST}
\begin{figure}[hb]
\includegraphics[width=\textwidth]{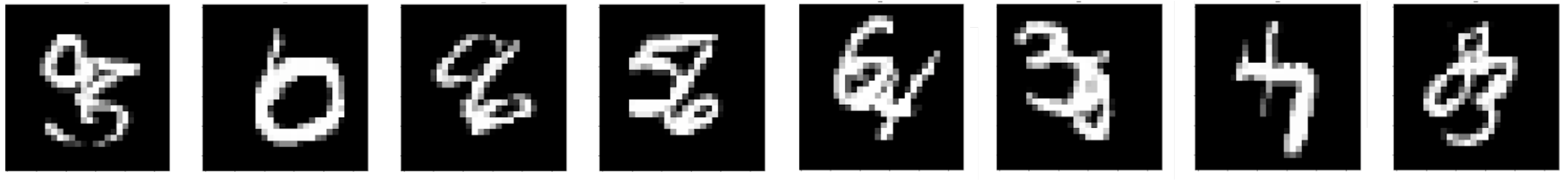}
\caption{Sample MultiMNIST images. In each image, one task (task-L) is classifying the digit on the top-left and the second task (task-R) is classifying the digit on the bottom-right.}
\label{fig:sample_multi_mnist}
\end{figure}

We use the MultiMNIST dataset, which overlays multiple images together \citep{multi_mnist}. For each image, a different one is chosen uniformly in random. One of these images is placed at the top-left and the other at the bottom-right. We show sample MultiMNIST images in Figure~\ref{fig:sample_multi_mnist}.

\begin{figure}[ht]
\includegraphics[width=\textwidth]{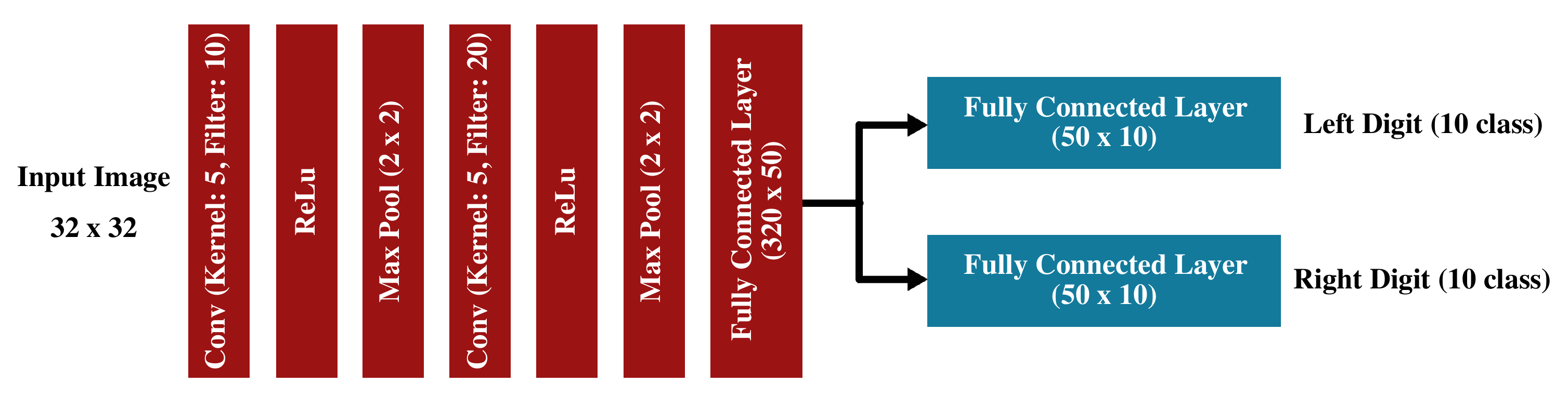}
\caption{Architecture used for MultiMNIST experiments.}
\label{fig:multi_mnist}
\end{figure}

For the MultiMNIST experiments, we use an architecture based on LeNet~\citep{mnist}. We use all layers except the final one as a shared encoder. We use the fully-connected layer as a task-specific function for the left and right tasks by simply adding two independent fully-connected layers, each taking the output of the shared encoder as input. As a task-specific loss function, we use the cross-entropy loss with a softmax for both tasks. The architecture is visualized in Figure~\ref{fig:multi_mnist}.

The implementation uses PyTorch \citep{pytorch}. For all baselines, we searched over the set $LR=\{\num{1e-4}, \num{5e-4}, \num{1e-3}, \num{5e-3}, \num{1e-2}, \num{5e-2}\}$ of learning rates and chose the model with the highest validation accuracy. We used SGD with momentum, halving the learning rate every 30 epochs. We use batch size $256$ and train for $100$ epochs. We report test accuracy.

\subsection{Multi-label classification}
\begin{figure}[ht]
\includegraphics[width=\textwidth]{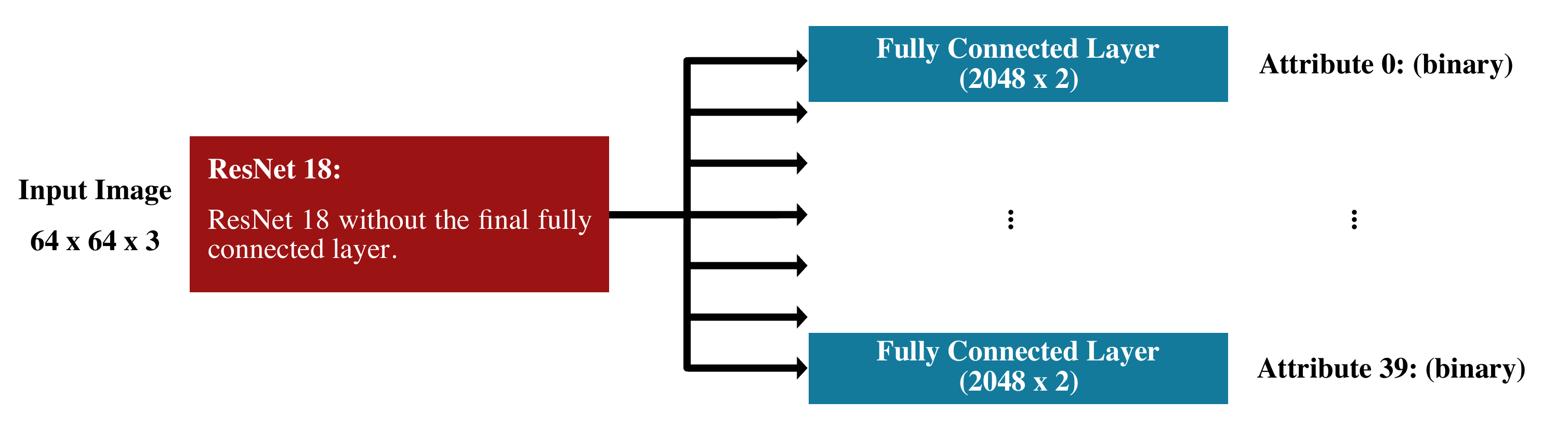}
\caption{Architecture used for multi-label classification experiments.}
\label{fig:arch_multi_label}
\end{figure}

For multi-label classification experiments, we use ResNet-18 \citep{resnet} without the final layer as a shared representation function. Since there are 40 attributes, we add 40 separate $2048\times2$ dimensional fully-connected layers as task-specific functions. The final two-dimensional output is passed through a 2-class softmax to get binary attribute classification probabilities. We use cross-entropy as a task-specific loss. The architecture is visualized in Figure~\ref{fig:arch_multi_label}.

The implementation uses PyTorch \citep{pytorch}. We resize each CelebA image \citep{celeba} to $64\times64\times3$. For all experiments, we searched over the set $LR=\{\num{1e-4}, \num{5e-4}, \num{1e-3}, \num{5e-3}, \num{1e-2}, \num{5e-2}\}$ of learning rates and chose the model with the highest validation accuracy. We used SGD with momentum, halving the learning rate every 30 epochs. We use batch size $256$ and train for $100$ epochs. We report attribute-wise binary accuracies on the test set as well as the average accuracy.

\subsection{Scene understanding}
\begin{figure}[ht]
\includegraphics[width=\textwidth]{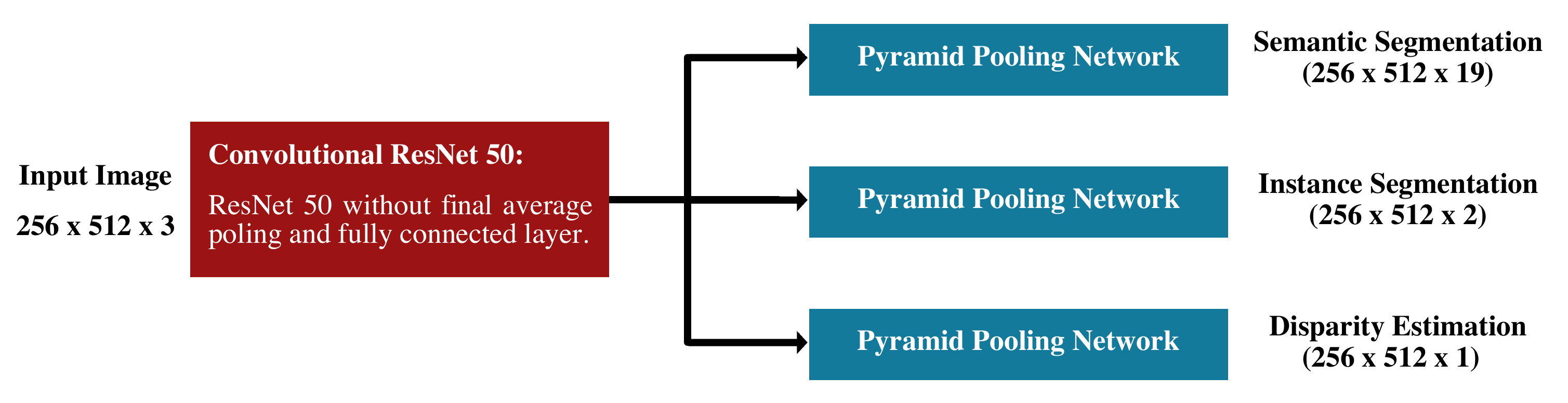}
\caption{Architecture used for scene understanding experiments.}
\label{fig:scene}
\end{figure}

For scene understanding experiments, we use the Cityscapes dataset \citep{cityscapes}. We resize all images to resolution $256\times512$ for computational efficiency. As a shared representation function (encoder), we use the ResNet-50 architecture \citep{resnet} in fully-convolutional fashion. We take the ResNet-50 architecture and only use layers prior to average pooling that are fully convolutional. As a decoder, we use the pyramid pooling module \citep{pspnet} and set the output sizes to $256\times512\times19$ for semantic segmentation ($19$ classes), $256\times512\times2$ for instance segmentation (one output channel for the x-offset of the center location and another channel for the y-offset), and $256\times512\times1$ for monocular depth estimation. For instance segmentation, we use the proxy task of estimating the offset for the center location of the instance that encompasses the pixel. We directly estimate disparity instead of depth and later convert it to depth using the provided camera intrinsics. As a loss function, we use cross-entropy with a softmax for semantic segmentation, and MSE for depth and instance segmentation. We visualize the architecture in Figure~\ref{fig:scene}.

We initialize the encoder with a model pretrained on ImageNet \citep{imagenet}. We use the implementation of the pyramidal pooling network with bilinear interpolation shared by \citet{pspnet_implementation}. Ground-truth results for the Cityscapes test set are not publicly available. Therefore, we report numbers on the validation set. As a validation set for hyperparameter search, we randomly choose $275$ images from the training set. After the best hyperparameters are chosen, we retrain with the full training set and report the metrics on the Cityscapes validation set, which our algorithm never sees during training or hyperparameter search. As metrics, we use mean intersection over union (mIoU) for semantic segmentation, MSE for instance segmentation, and MSE for disparities (depth estimation). We directly report the metric in the proxy task for instance segmentation instead of performing a further clustering operation. For all experiments, we searched over the set $LR=\{\num{1e-4}, \num{5e-4}, \num{1e-3}, \num{5e-3}, \num{1e-2}, \num{5e-2}\}$ of learning rates and chose the model with the highest validation accuracy. We used SGD with momentum, halving the learning rate every 30 epochs. We use batch size $8$ and train for $250$ epochs.

%% file: multi_task_arxiv.bbl
\clearpage

{\small


\begin{thebibliography}{60}
\providecommand{\natexlab}[1]{#1}
\providecommand{\url}[1]{\texttt{#1}}
\expandafter\ifx\csname urlstyle\endcsname\relax
  \providecommand{\doi}[1]{doi: #1}\else
  \providecommand{\doi}{doi: \begingroup \urlstyle{rm}\Url}\fi

\bibitem[Argyriou et~al.(2007)Argyriou, Evgeniou, and Pontil]{Argyriou2007}
A.~Argyriou, T.~Evgeniou, and M.~Pontil.
\newblock Multi-task feature learning.
\newblock In \emph{NIPS}, 2007.

\bibitem[Bagherjeiran et~al.(2005)Bagherjeiran, Vilalta, and
  Eick]{Bagherjeiran2005}
A.~Bagherjeiran, R.~Vilalta, and C.~F. Eick.
\newblock Content-based image retrieval through a multi-agent meta-learning
  framework.
\newblock In \emph{International Conference on Tools with Artificial
  Intelligence}, 2005.

\bibitem[Bakker and Heskes(2003)]{Bakker2003}
B.~Bakker and T.~Heskes.
\newblock Task clustering and gating for {Bayesian} multitask learning.
\newblock \emph{JMLR}, 4:\penalty0 83--99, 2003.

\bibitem[Baxter(2000)]{Baxter2000}
J.~Baxter.
\newblock A model of inductive bias learning.
\newblock \emph{Journal of Artificial Intelligence Research}, 12:\penalty0
  149--198, 2000.

\bibitem[Bilen and Vedaldi(2016)]{Bilen2016}
H.~Bilen and A.~Vedaldi.
\newblock Integrated perception with recurrent multi-task neural networks.
\newblock In \emph{NIPS}, 2016.

\bibitem[Caruana(1997)]{Caruana1997}
R.~Caruana.
\newblock Multitask learning.
\newblock \emph{Machine Learning}, 28\penalty0 (1):\penalty0 41--75, 1997.

\bibitem[Chen et~al.(2018)Chen, Badrinarayanan, Lee, and Rabinovich]{Chen2018}
Z.~Chen, V.~Badrinarayanan, C.~Lee, and A.~Rabinovich.
\newblock {GradNorm}: Gradient normalization for adaptive loss balancing in
  deep multitask networks.
\newblock In \emph{{ICML}}, 2018.

\bibitem[Collobert and Weston(2008)]{Collobert2008}
R.~Collobert and J.~Weston.
\newblock A unified architecture for natural language processing: Deep neural
  networks with multitask learning.
\newblock In \emph{ICML}, 2008.

\bibitem[Cordts et~al.(2016)Cordts, Omran, Ramos, Rehfeld, Enzweiler, Benenson,
  Franke, Roth, and Schiele]{cityscapes}
M.~Cordts, M.~Omran, S.~Ramos, T.~Rehfeld, M.~Enzweiler, R.~Benenson,
  U.~Franke, S.~Roth, and B.~Schiele.
\newblock The {Cityscapes} dataset for semantic urban scene understanding.
\newblock In \emph{{CVPR}}, 2016.

\bibitem[de~Miranda et~al.(2012)de~Miranda, Prud{\^{e}}ncio, de~Carvalho, and
  Soares]{Miranda2012}
P.~B.~C. de~Miranda, R.~B.~C. Prud{\^{e}}ncio, A.~C. P. L.~F. de~Carvalho, and
  C.~Soares.
\newblock Combining a multi-objective optimization approach with meta-learning
  for {SVM} parameter selection.
\newblock In \emph{International Conference on Systems, Man, and Cybernetics},
  2012.

\bibitem[Deng et~al.(2009)Deng, Dong, Socher, Li, Li, and Fei-Fei]{imagenet}
J.~Deng, W.~Dong, R.~Socher, L.-J. Li, K.~Li, and L.~Fei-Fei.
\newblock {ImageNet}: A large-scale hierarchical image database.
\newblock In \emph{{CVPR}}, 2009.

\bibitem[D{\'e}sid{\'e}ri(2012)]{Desideri2012}
J.-A. D{\'e}sid{\'e}ri.
\newblock Multiple-gradient descent algorithm {(MGDA)} for multiobjective
  optimization.
\newblock \emph{Comptes Rendus Mathematique}, 350\penalty0 (5):\penalty0
  313--318, 2012.

\bibitem[Dong et~al.(2015)Dong, Wu, He, Yu, and Wang]{Dong2015}
D.~Dong, H.~Wu, W.~He, D.~Yu, and H.~Wang.
\newblock Multi-task learning for multiple language translation.
\newblock In \emph{ACL}, 2015.

\bibitem[Ehrgott(2005)]{Ehrgott2005}
M.~Ehrgott.
\newblock \emph{Multicriteria Optimization {(2.} ed.)}.
\newblock Springer, 2005.

\bibitem[Fliege and Svaiter(2000)]{Fliege2000}
J.~Fliege and B.~F. Svaiter.
\newblock Steepest descent methods for multicriteria optimization.
\newblock \emph{Mathematical Methods of Operations Research}, 51\penalty0
  (3):\penalty0 479--494, 2000.

\bibitem[Ghosh et~al.(2013)Ghosh, Lovell, and Gunn]{Ghish2013}
S.~Ghosh, C.~Lovell, and S.~R. Gunn.
\newblock Towards {Pareto} descent directions in sampling experts for multiple
  tasks in an on-line learning paradigm.
\newblock In \emph{AAAI Spring Symposium: Lifelong Machine Learning}, 2013.

\bibitem[Hashimoto et~al.(2017)Hashimoto, Xiong, Tsuruoka, and
  Socher]{Hashimoto2016}
K.~Hashimoto, C.~Xiong, Y.~Tsuruoka, and R.~Socher.
\newblock A joint many-task model: Growing a neural network for multiple {NLP}
  tasks.
\newblock In \emph{EMNLP}, 2017.

\bibitem[He et~al.(2016)He, Zhang, Ren, and Sun]{resnet}
K.~He, X.~Zhang, S.~Ren, and J.~Sun.
\newblock Deep residual learning for image recognition.
\newblock In \emph{{CVPR}}, 2016.

\bibitem[Hern{\'{a}}ndez{-}Lobato et~al.(2016)Hern{\'{a}}ndez{-}Lobato,
  Hern{\'{a}}ndez{-}Lobato, Shah, and Adams]{Lobato2016}
D.~Hern{\'{a}}ndez{-}Lobato, J.~M. Hern{\'{a}}ndez{-}Lobato, A.~Shah, and R.~P.
  Adams.
\newblock Predictive entropy search for multi-objective bayesian optimization.
\newblock In \emph{{ICML}}, 2016.

\bibitem[Huang et~al.(2013)Huang, Li, Yu, Deng, and Gong]{Huang2013}
J.-T. Huang, J.~Li, D.~Yu, L.~Deng, and Y.~Gong.
\newblock Cross-language knowledge transfer using multilingual deep neural
  network with shared hidden layers.
\newblock In \emph{ICASSP}, 2013.

\bibitem[Huang et~al.(2015)Huang, Li, Siniscalchi, Chen, Wu, and
  Lee]{Huang2015}
Z.~Huang, J.~Li, S.~M. Siniscalchi, I.-F. Chen, J.~Wu, and C.-H. Lee.
\newblock Rapid adaptation for deep neural networks through multi-task
  learning.
\newblock In \emph{Interspeech}, 2015.

\bibitem[Jaggi(2013)]{Jaggi2013}
M.~Jaggi.
\newblock Revisiting {Frank-Wolfe}: Projection-free sparse convex optimization.
\newblock In \emph{{ICML}}, 2013.

\bibitem[Kaiser et~al.(2017)Kaiser, Gomez, Shazeer, Vaswani, Parmar, Jones, and
  Uszkoreit]{Kaiser2017}
L.~Kaiser, A.~N. Gomez, N.~Shazeer, A.~Vaswani, N.~Parmar, L.~Jones, and
  J.~Uszkoreit.
\newblock One model to learn them all.
\newblock \emph{arXiv:1706.05137}, 2017.

\bibitem[Kendall et~al.(2018)Kendall, Gal, and Cipolla]{Kendall2018}
A.~Kendall, Y.~Gal, and R.~Cipolla.
\newblock Multi-task learning using uncertainty to weigh losses for scene
  geometry and semantics.
\newblock In \emph{{CVPR}}, 2018.

\bibitem[Kokkinos(2017)]{Kokkinos2016}
I.~Kokkinos.
\newblock {UberNet}: Training a universal convolutional neural network for
  low-, mid-, and high-level vision using diverse datasets and limited memory.
\newblock In \emph{CVPR}, 2017.

\bibitem[Kuhn and Tucker(1951)]{Kuhn1951}
H.~W. Kuhn and A.~W. Tucker.
\newblock Nonlinear programming.
\newblock In \emph{Proceedings of the Second Berkeley Symposium on Mathematical
  Statistics and Probability}, Berkeley, Calif., 1951. University of California
  Press.

\bibitem[LeCun et~al.(1998)LeCun, Bottou, Bengio, and Haffner]{mnist}
Y.~LeCun, L.~Bottou, Y.~Bengio, and P.~Haffner.
\newblock Gradient-based learning applied to document recognition.
\newblock \emph{Proceedings of the IEEE}, 86\penalty0 (11):\penalty0
  2278--2324, 1998.

\bibitem[Li et~al.(2014)Li, Georgiopoulos, and Anagnostopoulos]{Cong2014}
C.~Li, M.~Georgiopoulos, and G.~C. Anagnostopoulos.
\newblock Pareto-path multi-task multiple kernel learning.
\newblock \emph{arXiv:1404.3190}, 2014.

\bibitem[Liu et~al.(2015{\natexlab{a}})Liu, Gao, He, Deng, Duh, and
  Wang]{Liu2015}
X.~Liu, J.~Gao, X.~He, L.~Deng, K.~Duh, and Y.-Y. Wang.
\newblock Representation learning using multi-task deep neural networks for
  semantic classification and information retrieval.
\newblock In \emph{NAACL HLT}, 2015{\natexlab{a}}.

\bibitem[Liu et~al.(2015{\natexlab{b}})Liu, Luo, Wang, and Tang]{celeba}
Z.~Liu, P.~Luo, X.~Wang, and X.~Tang.
\newblock Deep learning face attributes in the wild.
\newblock In \emph{{ICCV}}, 2015{\natexlab{b}}.

\bibitem[Long and Wang(2015)]{Long2015}
M.~Long and J.~Wang.
\newblock Learning multiple tasks with deep relationship networks.
\newblock \emph{arXiv:1506.02117}, 2015.

\bibitem[Luong et~al.(2015)Luong, Le, Sutskever, Vinyals, and
  Kaiser]{Luong2015}
M.-T. Luong, Q.~V. Le, I.~Sutskever, O.~Vinyals, and L.~Kaiser.
\newblock Multi-task sequence to sequence learning.
\newblock \emph{arXiv:1511.06114}, 2015.

\bibitem[Makimoto et~al.(1994)Makimoto, Nakagawa, and Tamura]{Makimoto1994}
N.~Makimoto, I.~Nakagawa, and A.~Tamura.
\newblock An efficient algorithm for finding the minimum norm point in the
  convex hull of a finite point set in the plane.
\newblock \emph{Operations Research Letters}, 16\penalty0 (1):\penalty0 33--40,
  1994.

\bibitem[Miettinen(1998)]{Miettinen1999}
K.~Miettinen.
\newblock \emph{Nonlinear Multiobjective Optimization}.
\newblock Springer, 1998.

\bibitem[Misra et~al.(2016)Misra, Shrivastava, Gupta, and Hebert]{Misra2016}
I.~Misra, A.~Shrivastava, A.~Gupta, and M.~Hebert.
\newblock Cross-stitch networks for multi-task learning.
\newblock In \emph{CVPR}, 2016.

\bibitem[Parisi et~al.(2014)Parisi, Pirotta, Smacchia, Bascetta, and
  Restelli]{Parisi2014}
S.~Parisi, M.~Pirotta, N.~Smacchia, L.~Bascetta, and M.~Restelli.
\newblock Policy gradient approaches for multi-objective sequential decision
  making.
\newblock In \emph{IJCNN}, 2014.

\bibitem[Paszke et~al.(2017)Paszke, Gross, Chintala, Chanan, Yang, DeVito, Lin,
  Desmaison, Antiga, and Lerer]{pytorch}
A.~Paszke, S.~Gross, S.~Chintala, G.~Chanan, E.~Yang, Z.~DeVito, Z.~Lin,
  A.~Desmaison, L.~Antiga, and A.~Lerer.
\newblock Automatic differentiation in {PyTorch}.
\newblock In \emph{NIPS Workshops}, 2017.

\bibitem[Peitz and Dellnitz(2018)]{Peitz2017}
S.~Peitz and M.~Dellnitz.
\newblock Gradient-based multiobjective optimization with uncertainties.
\newblock In \emph{NEO}, 2018.

\bibitem[Pirotta and Restelli(2016)]{Pirotta2016}
M.~Pirotta and M.~Restelli.
\newblock Inverse reinforcement learning through policy gradient minimization.
\newblock In \emph{AAAI}, 2016.

\bibitem[Poirion et~al.(2017)Poirion, Mercier, and
  D{\'{e}}sid{\'{e}}ri]{Poirion2017}
F.~Poirion, Q.~Mercier, and J.~D{\'{e}}sid{\'{e}}ri.
\newblock Descent algorithm for nonsmooth stochastic multiobjective
  optimization.
\newblock \emph{Computational Optimization and Applications}, 68\penalty0
  (2):\penalty0 317--331, 2017.

\bibitem[Roijers et~al.(2013)Roijers, Vamplew, Whiteson, and
  Dazeley]{Whiteson2018}
D.~M. Roijers, P.~Vamplew, S.~Whiteson, and R.~Dazeley.
\newblock A survey of multi-objective sequential decision-making.
\newblock \emph{Journal of Artificial Intelligence Research}, 48:\penalty0
  67--113, 2013.

\bibitem[Rosenbaum et~al.(2017)Rosenbaum, Klinger, and Riemer]{Rosenbaum2017}
C.~Rosenbaum, T.~Klinger, and M.~Riemer.
\newblock Routing networks: Adaptive selection of non-linear functions for
  multi-task learning.
\newblock \emph{arXiv:1711.01239}, 2017.

\bibitem[Rudd et~al.(2016)Rudd, G{\"u}nther, and Boult]{Rudd2016}
E.~M. Rudd, M.~G{\"u}nther, and T.~E. Boult.
\newblock {MOON}: A mixed objective optimization network for the recognition of
  facial attributes.
\newblock In \emph{ECCV}, 2016.

\bibitem[Ruder(2017)]{Ruder2017}
S.~Ruder.
\newblock An overview of multi-task learning in deep neural networks.
\newblock \emph{arXiv:1706.05098}, 2017.

\bibitem[Sabour et~al.(2017)Sabour, Frosst, and Hinton]{multi_mnist}
S.~Sabour, N.~Frosst, and G.~E. Hinton.
\newblock Dynamic routing between capsules.
\newblock In \emph{{NIPS}}, 2017.

\bibitem[Sch{\"a}ffler et~al.(2002)Sch{\"a}ffler, Schultz, and
  Weinzierl]{Schaffler2002}
S.~Sch{\"a}ffler, R.~Schultz, and K.~Weinzierl.
\newblock Stochastic method for the solution of unconstrained vector
  optimization problems.
\newblock \emph{Journal of Optimization Theory and Applications}, 114\penalty0
  (1):\penalty0 209--222, 2002.

\bibitem[Sekitani and Yamamoto(1993)]{Sekitani1993}
K.~Sekitani and Y.~Yamamoto.
\newblock A recursive algorithm for finding the minimum norm point in a
  polytope and a pair of closest points in two polytopes.
\newblock \emph{Mathematical Programming}, 61\penalty0 (1-3):\penalty0
  233--249, 1993.

\bibitem[Seltzer and Droppo(2013)]{Seltzer2013}
M.~L. Seltzer and J.~Droppo.
\newblock Multi-task learning in deep neural networks for improved phoneme
  recognition.
\newblock In \emph{ICASSP}, 2013.

\bibitem[Shah and Ghahramani(2016)]{Shah2016}
A.~Shah and Z.~Ghahramani.
\newblock Pareto frontier learning with expensive correlated objectives.
\newblock In \emph{{ICML}}, 2016.

\bibitem[Stein(1956)]{Stein1956}
C.~Stein.
\newblock Inadmissibility of the usual estimator for the mean of a multivariate
  normal distribution.
\newblock Technical report, Stanford University, US, 1956.

\bibitem[Wolfe(1976)]{Wolfe1976}
P.~Wolfe.
\newblock Finding the nearest point in a polytope.
\newblock \emph{Mathematical Programming}, 11\penalty0 (1):\penalty0 128--149,
  1976.

\bibitem[Xue et~al.(2007)Xue, Liao, Carin, and Krishnapuram]{Xue2007}
Y.~Xue, X.~Liao, L.~Carin, and B.~Krishnapuram.
\newblock Multi-task learning for classification with dirichlet process priors.
\newblock \emph{JMLR}, 8:\penalty0 35--63, 2007.

\bibitem[Yang and Hospedales(2016)]{Yang2017}
Y.~Yang and T.~M. Hospedales.
\newblock Trace norm regularised deep multi-task learning.
\newblock \emph{arXiv:1606.04038}, 2016.

\bibitem[Zamir et~al.(2018)Zamir, Sax, Shen, Guibas, Malik, and
  Savarese]{Zamir2018}
A.~R. Zamir, A.~Sax, W.~B. Shen, L.~J. Guibas, J.~Malik, and S.~Savarese.
\newblock Taskonomy: Disentangling task transfer learning.
\newblock In \emph{CVPR}, 2018.

\bibitem[Zhang and Yeung(2010)]{Zhang2010}
Y.~Zhang and D.~Yeung.
\newblock A convex formulation for learning task relationships in multi-task
  learning.
\newblock In \emph{{UAI}}, 2010.

\bibitem[Zhao et~al.(2017)Zhao, Shi, Qi, Wang, and Jia]{pspnet}
H.~Zhao, J.~Shi, X.~Qi, X.~Wang, and J.~Jia.
\newblock Pyramid scene parsing network.
\newblock In \emph{{CVPR}}, 2017.

\bibitem[Zhou et~al.(2017{\natexlab{a}})Zhou, Zhao, Puig, Fidler, Barriuso, and
  Torralba]{pspnet_implementation}
B.~Zhou, H.~Zhao, X.~Puig, S.~Fidler, A.~Barriuso, and A.~Torralba.
\newblock Scene parsing through {ADE20K} dataset.
\newblock In \emph{{CVPR}}, 2017{\natexlab{a}}.

\bibitem[Zhou et~al.(2017{\natexlab{b}})Zhou, Wang, Jiang, Guo, and
  Li]{ZhouDi2017}
D.~Zhou, J.~Wang, B.~Jiang, H.~Guo, and Y.~Li.
\newblock Multi-task multi-view learning based on cooperative multi-objective
  optimization.
\newblock \emph{IEEE Access}, 2017{\natexlab{b}}.

\bibitem[Zhou et~al.(2011{\natexlab{a}})Zhou, Chen, and Ye]{Zhou2011}
J.~Zhou, J.~Chen, and J.~Ye.
\newblock Clustered multi-task learning via alternating structure optimization.
\newblock In \emph{{NIPS}}, 2011{\natexlab{a}}.

\bibitem[Zhou et~al.(2011{\natexlab{b}})Zhou, Chen, and Ye]{zhou2011malsar}
J.~Zhou, J.~Chen, and J.~Ye.
\newblock {MALSAR}: Multi-task learning via structural regularization.
\newblock \emph{Arizona State University}, 2011{\natexlab{b}}.

\end{thebibliography}


}
